\documentclass{egpubl}
\usepackage{amsmath}
\usepackage{booktabs} 
\usepackage[most]{tcolorbox}
\usepackage{subcaption}

\ConferenceSubmission   
\usepackage[T1]{fontenc}
\usepackage{dfadobe}  
\usepackage{cite}  
\BibtexOrBiblatex
\electronicVersion
\PrintedOrElectronic
\ifpdf \usepackage{graphicx} \pdfcompresslevel=9
\else \usepackage[dvips]{graphicx} \fi

\usepackage{egweblnk} 
\title[Perceptually Grounded and Semantics-Aware Evaluation for Holistic Co-Speech Gesture Generation]%
      {Perceptually Grounded and Semantics-Aware Evaluation for Holistic Co-Speech Gesture Generation}

\author{
Nick Milkin$^{1,2}$\thanks{Equal Contribution},
Lanmiao Liu$^{1,2,3}$\footnotemark[1],
Esam Ghaleb$^{1,3}$,
Asli Ozyurek$^{1,3}$,
Zerrin Yumak$^{2}$\\
$^{1}$ Max Planck Institute for Psycholinguistics, Nijmegen, The Netherlands\\
$^{2}$ Utrecht University, Utrecht, The Netherlands\\
$^{3}$ Donders Institute for Brain, Cognition and Behaviour, Nijmegen, The Netherlands\\
\texttt{nick.milkin, lanmiao.liu, esam.ghaleb, asli.ozyurek@mpi.nl,}\\
\texttt{z.yumak@uu.nl}
}

\begin{document}


\maketitle
\thispagestyle{empty}
\begin{abstract}

Holistic and semantics-aware co-speech gesture generation has advanced rapidly, yet evaluation remains behind: objective metrics do not consistently reflect human perception, and semantic appropriateness remains difficult to quantify. We present a perceptually grounded and semantics-aware benchmark that combines standardized model comparison, human-centered metric validation, and fine-grained semantic evaluation. We first curate a list of 13 objective metrics covering different aspects, including distributional similarity, geometric fidelity, kinematic quality, cross-modal synchrony, and semantic appropriateness. For the semantic-appropriateness category, we propose a new metric, Semantic Gesture Preservation (SGP), which measures how far semantic gestures in the ground truth are preserved in the generated gestures. For this, we augment the BEAT2 dataset's annotations using a multi-modal LLM. We then conduct a perceptual study where 101 participants score generated gestures among five dimensions, including human-likeness, motion diversity, absence of animation errors, speech timing and content match. We systematically analyze objective metric--subjective score correlations. Unlike Semantic Score (SC), which shows no significant association with the evaluated perceptual dimensions, SGP is selectively aligned with speech-aware human judgments. We construct five target-specific composite metrics aligned with the subjective dimensions. These composites improve perceptual alignment across all five dimensions, with the largest gains for absence of animation errors and content match, indicating that complementary objective signals can better approximate human judgments than individual metrics alone. Overall, our results show that objective metrics require validation against subjective evaluations. Gesture quality assessment is complex and can benefit from semantics-aware and composite objective metrics.  We hope that these findings shed light to and form the basis for further fine-grained semantics-aware and composite metric assessments across a larger number of methods. Demos and qualitative illustrations are available on the anonymized project page: https://holisticgesturebenchmark.github.io/HolisticGestureBenchmark/. Code for individual and composite metrics and annotations for the dataset will be provided upon acceptance with a Github page.


\end{abstract}  
\section{Introduction}
Human communication is fundamentally multimodal \cite{HOLLER2019639}, where speech is linked to non-verbal behaviors that encode intent, emphasis, and discourse structure. As a cornerstone for developing virtual humans, co-speech gestures must exhibit deep behavioral coherence and social appropriateness rather than just visual fidelity \cite{Carrozino2020, RobotBehavior2012,ghaleb2026investigating,ghaleb2025see}. While early learning-based approaches targeted rhythmic, acoustics-driven upper-body motion \cite{habibie2021learning}, the field has recently undergone a paradigm shift toward a holistic synthesis that integrates the body, hands, and face while leveraging richer linguistic and semantic conditioning \cite{liu2024EMAGE}. However, as these generative architectures grow in complexity, the methodology for their evaluation has lagged. Co-speech gesture quality is inherently multifaceted, demanding a synergy of naturalness, diversity, temporal synchrony, and semantic relevance. Capturing these attributes remains an open challenge, leaving current practices reliant on fragmented and narrow evaluation protocols.

A fundamental question therefore persists regarding whether prevalent objective metrics designed to quantify motion fidelity, diversity, or temporal alignment \cite{liu2022beat, ao2023gesturediffuclip} truly align with human perception. Recent evidence suggests a significant divergence between numerical gains and subjective preferences \cite{nagy2024genealeaderboardextended}, indicating that current metrics often fail to capture the nuanced perceptual dimensions of high-quality gestures. This disconnect is most pronounced in semantic gesture evaluation: a gesture may be kinematically natural and temporally synchronized yet remain semantically vacuous or mismatched. Unlike rhythmic motion, semantic gestures are sparse, context-dependent, and exhibit a complex many-to-many mapping between meaning and form, rendering conventional statistical measures ill-equipped to capture these high-level relationships.

Reliable perceptual validation is further complicated by a systemic lack of standardization across the literature, where disparate datasets, preprocessing pipelines, rendering conditions, and subjective study designs have fostered a ``Wild West'' of evaluation in non-verbal behavior generation \cite{nagy2026towards, KaziWildWest}. In this environment, performance gains reported in individual studies are often not directly comparable, as apparent improvements may stem from idiosyncratic experimental settings rather than fundamental algorithmic advances, necessitating a transition toward evaluating representative models under a common protocol. Establishing such a standardized setting is therefore essential for determining whether objective metrics genuinely reflect human perception and for enabling controlled analysis of semantic gesture quality.

In this work, we propose a unified evaluation framework for semantics-aware holistic co-speech gesture generation bridging the gap between perceptual validation and objective metrics. We first conduct a systematic study of objective-subjective correlations drawing inspiration from multi-dimensional evaluation in other domains (i.e facial animation \cite{Delbosc2025} and general body motion \cite{Rekik_2026}). Rather than treating objective metrics as isolated scores, we analyze their perceptual relevance across five distinct dimensions and construct five target-specific composite metrics that combine complementary objective signals to predict human judgments of human-likeness, motion diversity, absence of animation errors, speech timing, and content match. To address the semantic gesture evaluation gap, we further introduce an LLM-assisted framework for fine-grained semantic annotation and derive a new semantics-aware metric. We empirically validate this metric against human judgments of content appropriateness, assessing its efficacy as an automated proxy for semantic performance. To support these analyses under controlled and directly comparable conditions, we establish a benchmark involving four state-of-the-art holistic gesture generation models: EMAGE \cite{liu2024EMAGE}, SemTalk \cite{zhang2025semtalk}, GestureLSM \cite{liu2025gesturelsm}, and SemConFlow \cite{liu2026SemConFlow}, all evaluated under a shared data and rendering pipeline. By integrating these components, we provide a rigorous foundation for measuring progress in semantics-aware holistic co-speech gesture generation, offering insights into the strengths and limitations of the current evaluative toolkit. Our primary contributions are:
\begin{itemize}
    \item \textbf{Perceptual Validation and Target-Specific Composite Metrics:} We systematically evaluate objective metrics against human judgments across five perceptual dimensions and construct five target-specific composite predictors that combine complementary objective signals to better approximate dimension-specific human ratings.

    \item \textbf{Semantics-Aware Gesture Evaluation Framework:} We establish a scalable paradigm for fine-grained semantic gesture annotation leveraging the cross-modal reasoning of LLMs, contributing an annotated corpus of semantic gesture descriptions to the community. Building upon this, we introduce the Semantic Gesture Preservation (SGP) metric, a new objective measure for semantic evaluation that is assessed against human perception.

    \item \textbf{A Semantics-aware Holistic Co-Speech Gesture Generation Benchmark:} We provide standardized objective metrics and subjective scores comparison of four representative generative models under a unified dataset, rendering pipeline, and evaluation protocol, enabling direct and fair performance analysis.
\end{itemize}

\section{Related Work}
\subsection{Co-speech gesture generation}

Co-speech gesture generation aims to synthesize human motion conditioned on multimodal signals such as speech and text, with the goal of producing behavior that is both perceptually plausible and communicatively appropriate \cite{nyatsanga2023comprehensive}. Early data-driven approaches mainly focused on generating rhythmic motion that followed the prosodic structure of speech \cite{habibie2021learning}, whereas later methods increasingly incorporated linguistic and semantic information to improve the correspondence between speech content and generated gestures. Representative directions include combining low-level acoustic features with higher-level speech representations \cite{chen2024diffsheg}, explicitly modeling semantically rich gestures separately from rhythmic motion \cite{zhi2023livelyspeaker}, retrieving semantically relevant motion from external databases \cite{zhang2023remodiffuse,mughal2025retrieving}, and optimizing semantic consistency between speech and motion during training \cite{liu2025semges} using contrastive learning \cite{liu2026SemConFlow}. These developments reflect a broader shift from generating motion that is primarily synchronized with speech toward gestures that also capture communicative meaning.

In parallel, advances in motion representation have enabled increasingly expressive and holistic generation. Discrete latent representations based on VQ-VAEs \cite{peng2021generating} and RQ-VAEs \cite{lee2022autoregressive} have been widely adopted to model complex motion distributions and provide stronger priors for full-body synthesis. EMAGE \cite{liu2024EMAGE} models different body components through compositional VQ-VAEs, while SemTalk \cite{zhang2025semtalk} employs separate residual-quantized representations for different body regions. Together with other recent systems such as GestureLSM \cite{liu2025gesturelsm} and TalkSHOW \cite{yi2023generating}, these methods extend co-speech gesture generation toward coordinated face, hand, and body motion under richer multi-modal conditioning \cite{liu2026SemConFlow}. Despite rapid progress, evaluation of co-speech gesture synthesis methods remains an open topic as these methods are aspiring to be more semantically-aware and holistic. Current methods do not specifically evaluate the semantic aspect in depth and rely on a set of common objective metrics and limited subjective evaluations.

\subsection{Human-Centered Validation of Objective Metrics}

Human evaluation remains the most direct way to assess the perceptual quality of generated gestures and is widely regarded as the gold standard in co-speech gesture generation \cite{nagy2024genealeaderboardextended}. However, subjective results are sensitive to study design, including question formulation, presented modalities, and the perceptual dimensions participants are asked to judge \cite{nagy2026towards, Wolfert_2022_evaluation_practices}. The GENEA evaluations sought to reduce such confounds by separating motion-quality judgments from speech-dependent appropriateness, for example through muted motion evaluation and mismatched speech--motion pairs \cite{nagy2024genealeaderboardextended, Kucherenko_2024}. Nevertheless, subjective protocols remain heterogeneous across studies and are costly to reproduce at scale, motivating the widespread use of automatic objective metrics.

Objective metrics provide efficient and reproducible estimates of properties such as motion fidelity, diversity, and speech--motion alignment, yet their correspondence with human perception remains uncertain. Large-scale evaluations have shown that correlations between commonly used objective metrics and subjective judgments are not always strong \cite{Kucherenko_2024,nagy2024genealeaderboardextended, Nyatsanga2023}. In other fields such as general body motion Rekik et al.~\cite{Rekik_2026} and facial animation Delbosc et al.~\cite{Delbosc2025} researchers suggested to use combined metrics as an alternative. These studies suggest that complementary metrics can provide more informative perceptual signals than individual scores alone. However, for the field of co-speech gesture generation the idea of combined metrics particularly for holistic and semantics-aware methods have not been explored. In this work, we first systematically analyze correlations between automatic metrics and human judgments for recent semantics-aware and holistic methods. We explicitly extend the current repertoire of objective metrics to better capture these aspects and correlate them with a carefully selected list of subjective scores. Based on these findings, we propose combined objective metrics that better correlates with subjective evaluations.

\subsection{Assessing Semantic Grounding in Co-Speech Gestures}

Multi-modal large language models (MLLMs) have recently emerged as scalable tools for semantic annotation and automatic evaluation. Their ability to jointly reason over visual and linguistic information has enabled applications beyond conventional recognition, including fine-grained multi-modal assessment and data annotation. Gemini -2.5\cite{comanici2025gemini} has been investigated as a general-purpose evaluator for vision-language tasks, showing promising agreement with human judgments across several evaluation settings \cite{tang2026chartmuseum}. Hu et al.\cite{hu2026multimodal} studies the use of MLLMs for scoring, pairwise comparison, and ranking across multiple tasks. Recent dataset construction pipelines have also exploited multi-modal models for scalable annotation; \cite{zhu2026mmr} uses Gemini-2.5 to generate fine-grained video reasonin MLLMs can provide rich semantic supervision without requiring every annotation to be manually specified. However, their potential for systematically annotating and evaluating the semantic content of co-speech gestures remains largely unexplored.

Existing semantics-aware evaluation approaches instead rely on predefined semantic annotations \cite{liu2022beat} and learned cross-modal representations \cite{ao2023gesturediffuclip,liu2025semges, paar2026duogesture,liu2023human}. Semantic Relevant Gesture Recall (SRGR) \cite{liu2022beat} weights pose accuracy according to semantic relevance annotations, emphasizing motion correctness at semantically important moments. Semantic Score (SC) \cite{ao2023gesturediffuclip} measures similarity between speech and generated motion in a learned gesture-text embedding space, while Emotional Accuracy (EA) \cite{qi2024emotiongesture} evaluates whether generated motion conveys a target affective category. Canonical Correlation Analysis (CCA) \cite{Kucherenko_2024,Thompson_1984} provides a more indirect measure by quantifying shared structure between generated and reference motion. Although these metrics extend evaluation beyond low-level motion statistics, they provide only partial access to whether a generated gesture is semantically present, what meaning it conveys, and whether that meaning is consistent with the accompanying speech. Overall, fine-grained semantic gesture evaluation remains constrained by limited scalable annotation and by metrics that only indirectly capture gesture meaning. We address this gap by leveraging large multimodal models for detailed semantic gesture annotation and, building on these annotations, introduce a semantics-aware metric that takes into account micro-level annotations.

\section{Methodology}

We curate a diverse set of objective metrics covering multiple aspects of motion and introduce a new semantics metric based on MLLM-derived fine-grained motion annotations. We evaluate recent holistic and semantics-aware gesture models, conduct a perceptual study extending existing subjective criteria with motion diversity, absence of animation errors, and content match, and analyze objective--subjective correlations to derive potential composite metrics.

\subsection{Benchmark Objective Metrics}
To quantify model performance, we organize objective metrics into a multidimensional taxonomy spanning distributional similarity, geometric fidelity, kinematic quality, cross-modal synchrony, and semantic appropriateness. The selected metrics operate across latent, distributional, and raw-signal levels, providing complementary measures of physical plausibility and communicative quality in co-speech gesture generation.

\subsubsection{Latent-space Distribution-Based Metrics}

\paragraph{Fr\'echet Gesture Distance (FGD)}

Fr\'echet Gesture Distance (FGD) \cite{Yoon_2020_FGD} measures the discrepancy between real and generated motion distributions in a learned latent feature space. Let $\boldsymbol{\mu}^{A},\boldsymbol{\Sigma}^{A}$ and $\boldsymbol{\mu}^{G},\boldsymbol{\Sigma}^{G}$ denote the corresponding means and covariances. FGD is defined as
\begin{equation}
\mathrm{FGD}
=
\left\|
\boldsymbol{\mu}^{A}
-
\boldsymbol{\mu}^{G}
\right\|_2^2
+
\operatorname{Tr}
\left(
\boldsymbol{\Sigma}^{A}
+
\boldsymbol{\Sigma}^{G}
-
2
\left(
\boldsymbol{\Sigma}^{A}
\boldsymbol{\Sigma}^{G}
\right)^{1/2}
\right).
\end{equation}
Lower values indicate closer agreement between the two motion distributions, while the metric depends on the learned feature encoder and does not directly capture speech--gesture semantics.

\paragraph{Density} Density \cite{naeem2020reliable} measures how well generated samples align with local regions of the real-data manifold. Unlike binary precision, it counts the number of reference neighborhoods containing each generated feature. Let $\{X_i\}_{i=1}^{N}$ and $\{Y_j\}_{j=1}^{M}$ denote reference and generated features, respectively, and let $\mathrm{NND}_k(X_i)$ be the distance from $X_i$ to its $k$-th nearest reference neighbor. Density is defined as \begin{equation} \mathrm{Density}=\frac{1}{kM}\sum_{j=1}^{M}\sum_{i=1}^{N}\mathbf{1}\left[Y_j\in B\left(X_i,\mathrm{NND}_k(X_i)\right)\right], \end{equation} where $B(\cdot)$ denotes the local reference neighborhood. Higher values indicate that generated motions more frequently fall within well-supported regions of the reference distribution.

\paragraph{Coverage}
Coverage \cite{naeem2020reliable} complements Density by measuring how extensively the generated samples cover the support of the reference motion distribution. Using the same $k$-nearest-neighbor neighborhoods defined for Density, Coverage computes the proportion of reference samples whose local neighborhood contains at least one generated sample:

\begin{equation}
\mathrm{Coverage}
=
\frac{1}{N}
\sum_{i=1}^{N}
\mathbf{1}
\left[
\exists j
\text{ such that }
Y_j \in
B\left(X_i,\mathrm{NND}_k(X_i)\right)
\right].
\end{equation}

Higher Coverage indicates that the generated samples represent a larger fraction of the real-motion manifold. Density and Coverage therefore provide complementary estimates of generation fidelity and support coverage in latent feature space.

\subsubsection{Geometric Fidelity Metrics}
\paragraph{Chamfer Distance} Chamfer Distance \cite{Rekik_2026} measures geometric similarity between generated and reference motions using their root-aligned vertex point clouds. Let $S_i^A$ and $S_i^G$ denote the reference and generated vertex sets at frame $i$. The symmetric distance over $n$ frames is \begin{multline} C(\{S_i^A\}_{i=1}^{n},\{S_i^G\}_{i=1}^{n})=\frac{1}{n}\sum_{i=1}^{n}\bigg(\frac{1}{|S_i^A|}\sum_{x\in S_i^A}\min_{y\in S_i^G}\|x-y\|_2^2\\+\frac{1}{|S_i^G|}\sum_{y\in S_i^G}\min_{x\in S_i^A}\|x-y\|_2^2\bigg). \end{multline} Lower values indicate closer average surface geometry, while primarily reflecting geometric fidelity rather than perceptual or semantic appropriateness.

\paragraph{Hausdorff Distance} Hausdorff Distance \cite{Hausdorff} measures the largest geometric discrepancy between generated and reference motion surfaces, making it more sensitive to local deviations than average-based distances. For reference and generated point sets $S_i^A$ and $S_i^G$ at frame $i$, it is defined as \begin{multline} H(\{S_i^A\}_{i=1}^{n},\{S_i^G\}_{i=1}^{n})=\frac{1}{n}\sum_{i=1}^{n}\max\bigg\{\max_{x\in S_i^A}\min_{y\in S_i^G}\|x-y\|_2,\\ \max_{y\in S_i^G}\min_{x\in S_i^A}\|x-y\|_2\bigg\}. \end{multline} Lower values indicate smaller worst-case geometric deviations from the reference motion.

\paragraph{Mean Joint Distance} Mean Joint Distance (MJD) \cite{Rekik_2026} measures pose-space error between generated and reference motion using corresponding root-aligned body and hand joints. Let $p_{ct}^{j}$ and $\hat{p}_{cmt}^{j}$ denote the reference and generated positions of joint $j$ at frame $t$. MJD is defined as \begin{equation} \mathrm{MJD}_{cm}=\frac{1}{TJ}\sum_{t=1}^{T}\sum_{j=1}^{J}\left\|p_{ct}^{j}-\hat{p}_{cmt}^{j}\right\|_2, \end{equation} where $T$ and $J$ denote the numbers of aligned frames and evaluated joints, respectively. Lower values indicate closer agreement with the reference pose sequence.

\paragraph{Dice Coefficient}
The Dice Coefficient \cite{DiceCoefficient} measures spatial overlap between generated and reference gesture trajectories after projection onto a binary occupancy map in the frontal $xy$ plane. Let $A$ and $B$ denote the occupied regions of the reference and generated motions, respectively. It is defined as
\begin{equation}
\mathrm{Dice}(A,B)
=
\frac{2|A\cap B|}
{|A|+|B|}.
\end{equation}
The score lies in $[0,1]$, with higher values indicating greater spatial overlap. Dice captures the overall motion footprint but discards depth information and temporal ordering.

\subsubsection{Motion Quality and Physical Plausibility Metrics}

\paragraph{Foot Contact}
Foot Contact measures lower-body plausibility by comparing stable ground-contact rates between generated and reference motion. A foot is considered in contact when at least two of the heel, big-toe, and small-toe vertices are within $3$\,cm of the estimated floor and have horizontal velocity below $0.10$\,m/s. Let $r_{\mathrm{GT}}$ and $r_{\mathrm{GEN}}$ denote the proportions of frames with at least one foot in contact. The Foot Contact error is
\begin{equation}
\mathrm{FC}
=
\left|
r_{\mathrm{GT}}
-
r_{\mathrm{GEN}}
\right|.
\end{equation}
Lower values indicate closer agreement with reference contact behavior. The metric primarily reflects lower-body stability and artifacts such as implausible contact or foot sliding, rather than speech-related gesture quality.

\paragraph{Log Dimensionless Jerk (LDLJ)}
Log Dimensionless Jerk (LDLJ) \cite{motion_smoothness_ldlj} measures motion smoothness from the third derivative of position while normalizing for movement duration and amplitude. For a trajectory $p(t)$ with duration $T$ and path length $L$, it is defined as
\begin{equation}
\mathrm{LDLJ}(p)
=
-\ln
\left(
\frac{T^5}{L^2}
\int_{0}^{T}
\left(
\frac{d^3p(t)}{dt^3}
\right)^2
dt
\right).
\end{equation}
Higher LDLJ indicates smoother motion. To compare generated and reference sequences, we use the relative score $\mathrm{LDLJ}_{\mathrm{REL}}=\mathrm{LDLJ}_{\mathrm{GT}}-\mathrm{LDLJ}_{\mathrm{GEN}}$, where values closer to zero indicate more similar smoothness and positive values indicate smoother reference motion.

\subsubsection{Speech-Motion Alignment Metrics}
\paragraph{Beat Consistency}
Beat Consistency (BC) \cite{li2021ai} measures temporal synchronization between kinematic motion beats and acoustic beats in the accompanying speech. Let $B^x=\{t_i^x\}$ denote detected gesture-beat timestamps and $B^y=\{t_j^y\}$ detected audio-beat timestamps. BC is computed as

\begin{equation}
\mathrm{BC}
=
\frac{1}{m}
\sum_{i=1}^{m}
\exp
\left(
-
\frac{
\min_{t_j^y\in B^y}
\|t_i^x-t_j^y\|^2
}
{2\sigma^2}
\right),
\end{equation}

where $\sigma$ controls the temporal tolerance around each acoustic beat. Higher values indicate stronger temporal correspondence between gesture and speech beats. BC is therefore prosody-aware but semantics-agnostic: it evaluates \emph{when} a gesture occurs, but not whether the gesture expresses information consistent with the spoken content.

\subsubsection{Semantics-Aware Metrics}
\label{sec:Semantics-Aware Metrics}
\paragraph{SRGR}
Semantic Relevant Gesture Recall (SRGR) \cite{liu2022beat} extends PCK by weighting pose correctness according to semantic relevance. Let $p_t^j$ and $\hat{p}_t^j$ denote the reference and generated positions of joint $j$ at frame $t$, respectively. SRGR is defined as
\begin{equation}
D_{\mathrm{SRGR}}
=
\lambda
\frac{1}{TJ}
\sum_{t=1}^{T}
\sum_{j=1}^{J}
\mathbf{1}
\left[
\|p_t^j-\hat{p}_t^j\|_2<\delta
\right],
\end{equation}
where $\lambda$ is the semantic relevance weight and $\delta$ the distance threshold. Higher values indicate greater pose agreement at semantically important intervals. As a reference-pose-based metric, SRGR may penalize alternative but semantically valid gesture realizations.

\paragraph{Semantic Score}
Semantic Score (SC) \cite{ao2023gesturediffuclip} measures speech--gesture semantic alignment in a learned cross-modal embedding space by comparing transcript and generated-motion embeddings with cosine similarity. Let $\mathbf{e}_{\mathrm{text}}$ and $\mathbf{e}_{\mathrm{motion}}$ denote the corresponding embeddings. SC is defined as \begin{equation} \mathrm{SC}=\frac{\mathbf{e}_{\mathrm{text}}^{\top}\mathbf{e}_{\mathrm{motion}}}{\|\mathbf{e}_{\mathrm{text}}\|_2\|\mathbf{e}_{\mathrm{motion}}\|_2}. \end{equation} Higher values indicate stronger semantic alignment. Unlike SRGR, SC does not require frame-level semantic annotations, but depends on the quality of the learned gesture--text embedding space.

\paragraph{Semantic Gesture Preservation (SGP)}
We introduce \textbf{Semantic Gesture Preservation (SGP)}, a new semantics-aware metric designed to evaluate whether generated motion preserves semantic gesturing during speech intervals annotated as semantically meaningful; see details in \ref{sec:LLM-based Semantic Gesture Annotation}. In contrast to reference-pose metrics such as SRGR, SGP does not require the generated gesture to reproduce the exact physical realization observed in the ground truth. This is important for co-speech gestures, where the same communicative meaning can be expressed through multiple valid motion forms.

SGP operates on semantic gesture events annotated in BEAT2 and uses Gemini~2.5~Pro \cite{comanici2025gemini} as a multimodal gesture classifier. For each annotated semantic event, the corresponding generated video is extracted with temporal context and presented to Gemini together with the speech and annotation metadata. The model is instructed to determine the visibly observed gesture rather than assume that a semantic gesture is present, and classifies its type as \textit{iconic}, \textit{metaphoric}, \textit{deictic}, \textit{beat}, \textit{none}, or \textit{uncertain}. This yields a scalable event-level assessment of whether intervals expected to contain semantic gestures instead collapse into generic beat-like motion.

Let $E_c$ denote the set of annotated semantic gesture events in clip $c$, and let $\hat{\tau}_{cme}$ denote the gesture type identified by Gemini for event $e$ from model $m$. An event is considered semantically degraded when $\hat{\tau}_{cme}=\mathrm{beat}$. The clip-level Semantic Gesture Preservation score is then defined as

\begin{equation}
\mathrm{SGP}_{cm}
=
1-
\frac{
n\left(
\hat{\tau}_{cme}=\mathrm{beat}
\right)
}{
n(E_c)
},
\end{equation}

where $n(E_c)$ is the number of valid semantic events in the reference clip and
$n(\hat{\tau}_{cme}=\mathrm{beat})$ counts how many of these intervals are classified as beat gestures in the generated motion. Higher values indicate that semantic-gesture intervals are less frequently replaced by generic beat gestures, with $\mathrm{SGP}=1$ indicating that none of the evaluated semantic events are classified as beats.

The model-level SGP is obtained by averaging the clip-level scores:

\begin{equation}
\mathrm{SGP}_{m}
=
\frac{1}{C_m}
\sum_{c=1}^{C_m}
\mathrm{SGP}_{cm},
\end{equation}

where $C_m$ denotes the number of evaluated clips for model $m$. Each clip therefore contributes equally regardless of its number of annotated semantic events.

SGP should be interpreted specifically as a measure of \emph{semantic gesture preservation against beat-gesture collapse}. It does not aim to verify that the generated gesture reproduces the exact annotated meaning or specific semantic gesture type. Instead, it provides a complementary semantic metric that shows how far semantic gestures in the reference motion are preserved in the generated motion.

\subsection{LLM-Based Semantic Gesture Annotation}
\label{sec:LLM-based Semantic Gesture Annotation}
The existing BEAT2 annotations identify semantically relevant gesture intervals and provide coarse gesture categories, but they contain limited information about how a gesture is physically realized or what communicative meaning it expresses in context. Such information is important for semantic evaluation because semantically equivalent gestures may differ substantially in their exact pose trajectories. We therefore enrich the existing intervals with structured multimodal annotations using Gemini~2.5~Pro \cite{comanici2025gemini}. For each predefined semantic interval \(e\), we extract the corresponding rendered motion segment and pair it with the annotated target word or expression. Gemini is instructed to analyze only the target interval and to ground its output in the visible motion. The annotation decomposes semantic gesture information into three complementary components: (i) \emph{gesture form}, describing handedness, hand shape, orientation, spatial location, and movement trajectory; (ii) \emph{contextual meaning}, describing the communicative interpretation of the observed gesture in relation to the accompanying expression; and (iii) \emph{gesture type}, selected from \{\textit{iconic}, \textit{metaphoric}, \textit{deictic}, \textit{beat}\}. The model additionally reports a confidence value. We represent the resulting annotation as
\begin{equation}
\mathcal{A}_e=\left(w_e,d_e,m_e,\tau_e,c_e\right),
\label{eq:semantic_annotation}
\end{equation}
where \(w_e\) denotes the target word or expression, \(d_e\) the fine-grained description of observable gesture form, \(m_e\) the inferred contextual meaning, \(\tau_e\) the gesture type, and \(c_e\) the annotation confidence. For example, for an interval associated with the word \emph{had}, the annotator describes a small downward right-hand motion in front of the torso, interprets the motion as conveying possession, and assigns the gesture to the \emph{iconic} category. This structured representation separates what is visibly performed from how that motion is interpreted, which is useful when multiple physical realizations can convey similar communicative content. The complete annotation prompt and additional examples are provided in Appendix~\ref{sec:annotation_procedure}. To assess annotation reliability, a subset of the generated annotations is manually evaluated along the same information dimensions used downstream: gesture-type agreement, contextual-meaning correctness, and the quality of the gesture-form description (Appendix~\ref{app:annotation_validation}). This validation does not treat Gemini outputs as semantic ground truth; rather, it assesses whether the generated annotations are sufficiently consistent and interpretable to support subsequent semantics-aware analysis. The enriched annotations are subsequently used in the Semantic Gesture Preservation metric described in Section~\ref{sec:Semantics-Aware Metrics}.
\subsection{Benchmark Models}
This subsection summarizes the representative holistic co-speech gesture generation models selected for our benchmark, covering diverse methodological paradigms within a unified full-body motion space. Architectural details are provided in the respective original papers.

\subsubsection{EMAGE} EMAGE \cite{liu2024EMAGE} proposes a unified framework for holistic co-speech gesture generation, jointly synthesizing facial expressions, upper- and lower-body motion, hand gestures, and global translation. The model uses a Masked Audio Gesture Transformer to jointly optimize masked gesture reconstruction and audio-conditioned generation, allowing partially observed motion to provide structured body hints during synthesis. To capture the heterogeneous dynamics of different body regions, EMAGE employs four compositional VQ-VAEs that separately model the face, upper body, hands, and lower body in discrete latent spaces. It further introduces Content Rhythm Attention (CRA), which adaptively combines rhythmic speech cues with transcript semantics to balance beat-aligned and content-related gesture generation. The reconstructed latent features are decoded into local facial and body motion, while a separate global motion predictor estimates body translation. This design enables coherent, audio-synchronized holistic gesture generation while supporting partially predefined spatial-temporal gesture constraints.

\subsubsection{GestureLSM} GestureLSM \cite{liu2025gesturelsm} introduces a flow-matching-based framework for real-time holistic co-speech gesture generation, with emphasis on modeling interactions across body regions. Rather than treating the face, hands, upper body, and lower body as independent streams, it represents them with separate residual vector-quantized latent tokens and captures their dependencies through sequential spatial and temporal attention. Spatial attention models inter-region coordination within each frame, while temporal attention captures motion evolution over time. Speech conditioning combines low-level acoustic cues with high-level semantic transcript features and is fused with gesture representations through cross-attention. To improve sampling efficiency, GestureLSM further introduces a latent shortcut model for learning direct transitions along the flow trajectory, together with beta-distributed timestep sampling. This design substantially reduces inference steps while maintaining high-quality, speech-coordinated holistic motion generation.

\subsubsection{SemTalk} SemTalk \cite{zhang2025semtalk} generates holistic co-speech motion by separating rhythm-related base motion from sparse semantic motion and adaptively integrating them at the frame level. A hierarchical coarse-to-fine cross-attention module propagates speech-driven motion cues from the face to the hands, upper body, and lower body, while local and global rhythm objectives encourage temporal synchronization and coherence. To model infrequent but semantically salient gestures, SemTalk introduces semantic emphasis learning that combines local text features, sentence-level CLIP representations, emotional cues, and high-level speech features to estimate frame-level semantic scores. These scores selectively activate semantic motion and control its fusion with the rhythm-aligned base motion. Separate RVQ-VAE representations are used for different body regions to preserve region-specific dynamics and support expressive holistic synthesis.

\subsubsection{SemConFlow} SemConFlow \cite{liu2026SemConFlow} introduces a semantically grounded framework for holistic co-speech gesture generation by combining multimodal semantic alignment with contrastive flow matching. It first learns region-specific motion priors for the face, hands, upper body, and lower body using hierarchical RVQ-VAEs and aggregates them into a composite full-body latent space. A Semantics-Aware Composite Module aligns this representation with textual and acoustic features through cosine and contrastive objectives, encouraging consistent semantics across body regions. SemConFlow then applies contrastive flow matching, using mismatched audio--text conditions as negative contexts so that the learned velocity field favors semantically congruent speech--motion trajectories while diverging from incongruent alternatives. Temporal cross-attention and self-attention further integrate multimodal conditioning and maintain motion coherence. Experiments on BEAT2 and SHOW demonstrate strong performance in motion realism, diversity, synchronization, and semantic consistency.

\subsection{Perception Study Design}
\begin{figure}[t]
    \centering
    \begin{minipage}{0.48\linewidth}
        \centering
        \includegraphics[width=\linewidth]{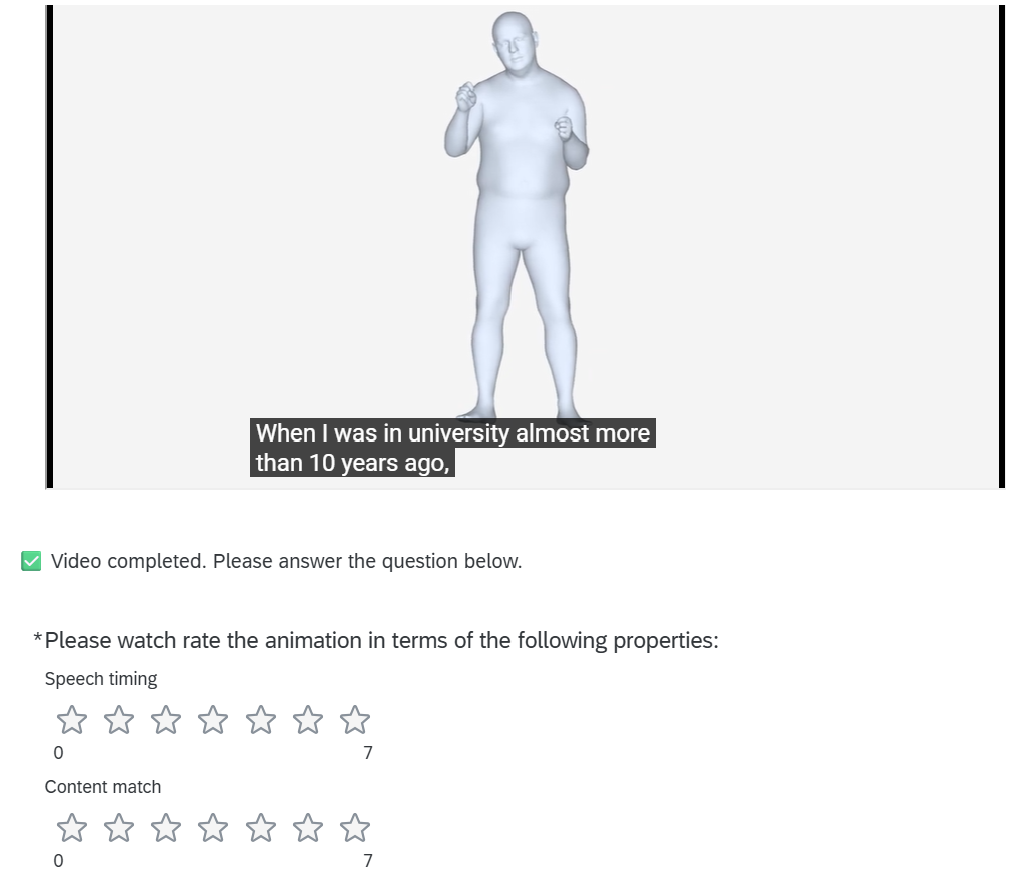}
    \end{minipage}
    \hfill
    \begin{minipage}{0.48\linewidth}
        \centering
        \includegraphics[width=\linewidth]{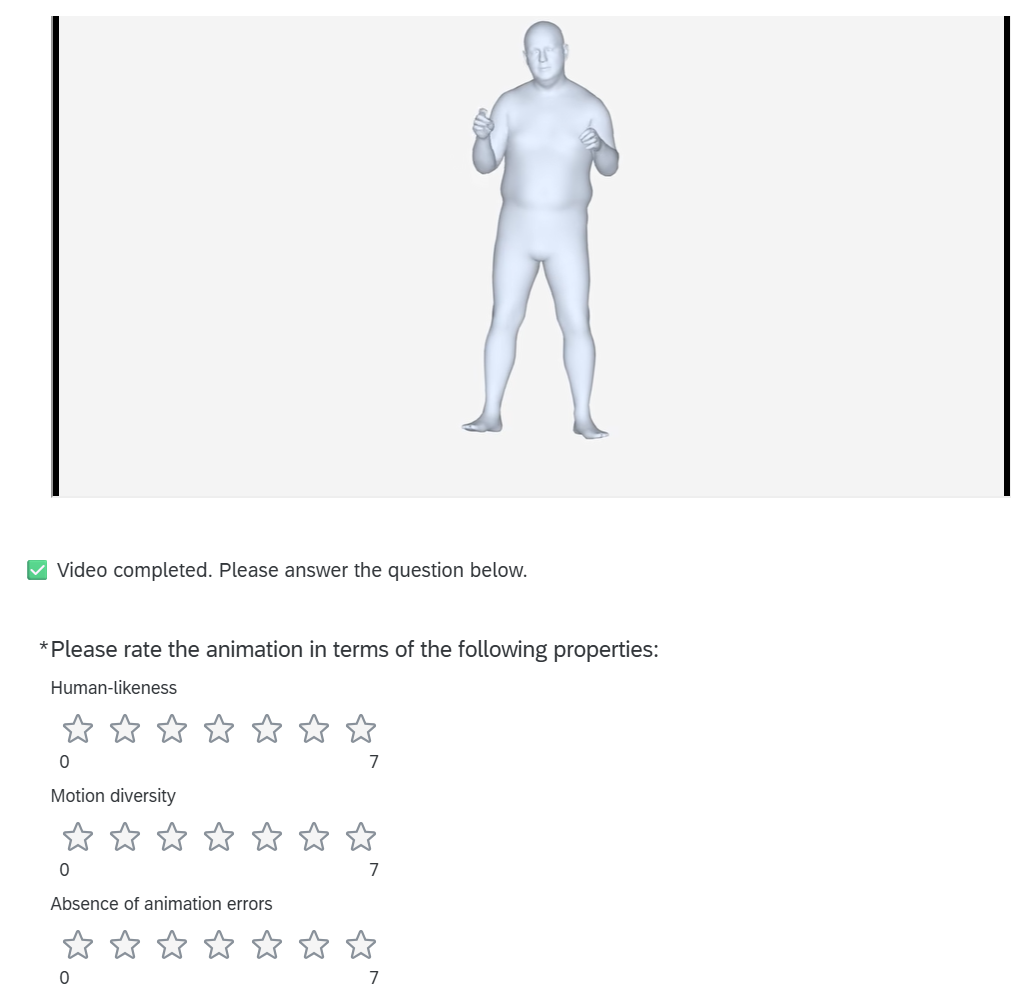}
    \end{minipage}
    \caption{
Perceptual study interfaces for the audio (left) and muted (right) conditions, evaluating speech-aware and visual-quality dimensions, respectively. Participants complete one condition using the same seven-point Likert scale and video interface.
    }
    \label{fig:perceptual_study_interface}
\end{figure}

We conduct a perceptual study to obtain human judgments of holistic co-speech gesture quality and use these ratings as targets for validating the objective metrics. Following the general separation adopted in GENEA-style evaluation \cite{nagy2026towards}, we distinguish visual motion quality from speech-dependent appropriateness through two independent conditions. Unlike GENEA, however, our perceptual criteria are selected to align with the multidimensional evaluation objectives of this benchmark. The muted condition assesses human-likeness, motion diversity, and absence of animation errors, capturing complementary aspects of visual motion quality. The audio condition evaluates speech timing and content match, explicitly separating temporal coordination from semantic correspondence. This distinction enables a more fine-grained analysis of how different objective metrics relate to specific perceptual properties of generated gestures. All criteria are rated on a seven-point Likert scale, with higher scores indicating better perceived quality. The two evaluation interfaces are shown in Figure~\ref{fig:perceptual_study_interface}.

The study compares EMAGE, SemConFlow, SemTalk, and GestureLSM together with BEAT2 motion-capture ground truth. All conditions use the same avatar, camera, crop, and rendering pipeline. Each stimulus is a 10-second clip. As we are specifically interested in evaluating semantically rich gesture clips, BEAT2 annotations are used to select windows with the highest occurrence of iconic, metaphoric, and deictic gestures. The same temporal interval is used for all models and the corresponding ground-truth motion. The selected stimuli contain 80 annotated semantic events: 44 iconic, 18 metaphoric, and 18 deictic gestures. A total of 103 participants were recruited online with compensation: 51 for the muted condition and 52 for the audio condition. After quality control, two participants were excluded from the muted study, resulting in 101 valid participants in total (49 muted and 52 audio).

Each study evaluates 15 unique sequences from five motion sources (four models and ground truth), yielding 75 unique source sequence pairs for analysis.Videos must be watched in full before ratings can be submitted, and participants failing two or more attention checks are excluded. For each model sequence pair, ratings are summarized by their mean and standard deviation, with the mean used as the subjective target in the objective--subjective correlation analysis. This yields 75 observations across all five motion sources and 60 generated-motion observations after excluding BEAT2 ground truth. We further compute pairwise Spearman correlations among the five subjective dimensions to assess the extent to which the selected perceptual criteria provide complementary judgments (Section \ref{subsec: relamongperceptual}). Further details of the user-study procedure and interface are provided in Appendix~\ref{sec:user_study_interface}.

\section{Experiments}
\subsection{Datasets}
BEAT2 \cite{liu2024EMAGE} contains approximately 60 hours of synchronized speech and holistic motion from 25 speakers (12 female and 13 male). The dataset is divided into BEAT2-Standard (27 hours) and BEAT2-Additional (33 hours), with the latter capturing more spontaneous and natural gestural behavior. Full-body motion is represented using SMPL-X \cite{pavlakos2019expressive} and further refined with MoSh++ \cite{mahmood2019amass}, providing detailed body, hand, facial, and global motion aligned with speech. Compared with the original BEAT dataset, BEAT2 provides a more accurate mesh-based representation with improved body proportions, hand articulation, and facial motion, making it well suited for holistic co-speech gesture generation and evaluation. In our experiments, we use the BEAT2-Standard subset and adopt the official 85\%/7.5\%/7.5\% train/validation/test split over all 25 speakers.
\subsection{Implementation Detail}
All benchmark experiments are conducted on a single NVIDIA A100 GPU under a unified evaluation setup. To ensure a consistent comparison, all evaluated methods are retrained using their official implementations and original optimization settings, while sharing the full BEAT2 training set as the common training-data regime. This extends the original single-speaker training setup used by EMAGE\cite{liu2024EMAGE}, SemTalk\cite{zhang2025semtalk}, and GestureLSM\cite{liu2025gesturelsm} to 60 hours of multi-speaker training data. During inference, all models are evaluated on the same BEAT2 test sequences with identical speech inputs and sequence boundaries. Their outputs are converted to a common SMPL-X representation and processed through the same post-processing, rendering, and metric-computation pipeline, reducing variation introduced by differences in implementation or evaluation protocol.

\subsection{Objective Evaluation}
Table~\ref{tab:model_level_objective} reveals pronounced differences across evaluation dimensions. SemConFlow achieves the strongest performance on most distributional, geometric, and speech-alignment metrics, including FGD, BC, Diversity, Dice, SRGR, PCK, MJD, Chamfer, Hausdorff, and SC. Its substantially lower FGD together with higher Diversity suggests a closer match to the reference motion distribution without sacrificing motion variability, while the improvements in MJD, Chamfer, and Hausdorff indicate stronger geometric agreement with the reference motion. SemConFlow also obtains the highest BC and SC, suggesting stronger temporal coordination with speech and higher cross-modal semantic similarity in the learned gesture--text embedding space.

No single method, however, dominates across all metric families. EMAGE achieves the best Density and Coverage and the relative LDLJ value closest to zero, indicating stronger support within the reference feature distribution and smoothness closer to the captured motion. Notably, EMAGE also obtains the highest SGP despite not achieving the highest SC. This discrepancy highlights an important distinction between the two semantics-aware measures: SC evaluates global speech--motion similarity in a learned embedding space, whereas SGP targets whether annotated semantic-gesture intervals are preserved rather than collapsing into generic beat-like motion. SemTalk performs best on Foot Contact, indicating more consistent lower-body contact behavior, while GestureLSM does not lead on an individual metric.

Overall, the varying model rankings across distributional, geometric, kinematic, temporal, and semantic metrics show that these measures capture complementary aspects of gesture quality. While SC and SGP scores overall indicates EMAGE and SemConFlow as best, overall rankings are different suggesting more analysis in the semantics-aware metrics direction. These results therefore motivate a multidimensional evaluation protocol and, more importantly, the subsequent analysis of which objective signals are actually aligned with human perception.

\begin{table*}[t]
\centering
\small
\setlength{\tabcolsep}{1.6pt}
\renewcommand{\arraystretch}{1.05}

\begin{tabular}{lccccccccccccccc}
\toprule
Method
& FGD$\downarrow$
& BC$\uparrow$
& Div.$\uparrow$
& Dens.$\uparrow$
& Cov.$\uparrow$
& Dice$\uparrow$
& LDLJ$_{\mathrm{REL}}\!\rightarrow\!0$
& SRGR$\uparrow$
& PCK$\uparrow$
& MJD$\downarrow$
& Chamfer$\downarrow$
& Hausdorff$\downarrow$
& Foot Contact$\downarrow$
& SC$\uparrow$
& SGP$\uparrow$ \\
\midrule

EMAGE
& 5.523
& 0.692
& 88
& \textbf{0.8972}
& \textbf{0.07275}
& 0.6407
& \textbf{0.0530}
& 0.05494
& 0.3309
& 0.2410
& 0.01375
& 0.2908
& 0.5048
& 0.248
& \textbf{0.4372} \\

SemConFlow
& \textbf{2.245}
& \textbf{0.780}
& \textbf{120}
& 0.01559
& 0.06647
& \textbf{0.6793}
& -0.4294
& \textbf{0.05868}
& \textbf{0.3552}
& \textbf{0.2007}
& \textbf{0.01112}
& \textbf{0.2530}
& 0.5932
& \textbf{0.314}
& 0.4205 \\

SemTalk
& 4.266
& 0.727
& 116
& 0.1123
& 0.06673
& 0.6241
& -0.6020
& 0.05059
& 0.3037
& 0.2303
& 0.13976
& 0.4663
& \textbf{0.4040}
& 0.268
& 0.4014 \\

GestureLSM
& 4.268
& 0.525
& 112
& 0.009239
& 0.009817
& 0.6573
& 1.5386
& 0.04536
& 0.2715
& 0.2391
& 0.01797
& 0.2909
& 0.8883
& 0.248
& 0.2622 \\
\bottomrule
\end{tabular}

\caption{
Objective comparison of the benchmarked holistic co-speech gesture generation methods on BEAT2. All metrics are computed over the 265 shared test sequences, except SGP, which is evaluated on the 15 sequences used in the perceptual study with Gemini-based semantic annotations. Arrows indicate the preferred direction, and bold denotes the best result for each metric.
}
\label{tab:model_level_objective}
\end{table*}

\subsection{Subjective Evaluation}
\subsubsection{Main Analysis.}
\begin{figure}[t]
    \centering
    \includegraphics[width=\linewidth]{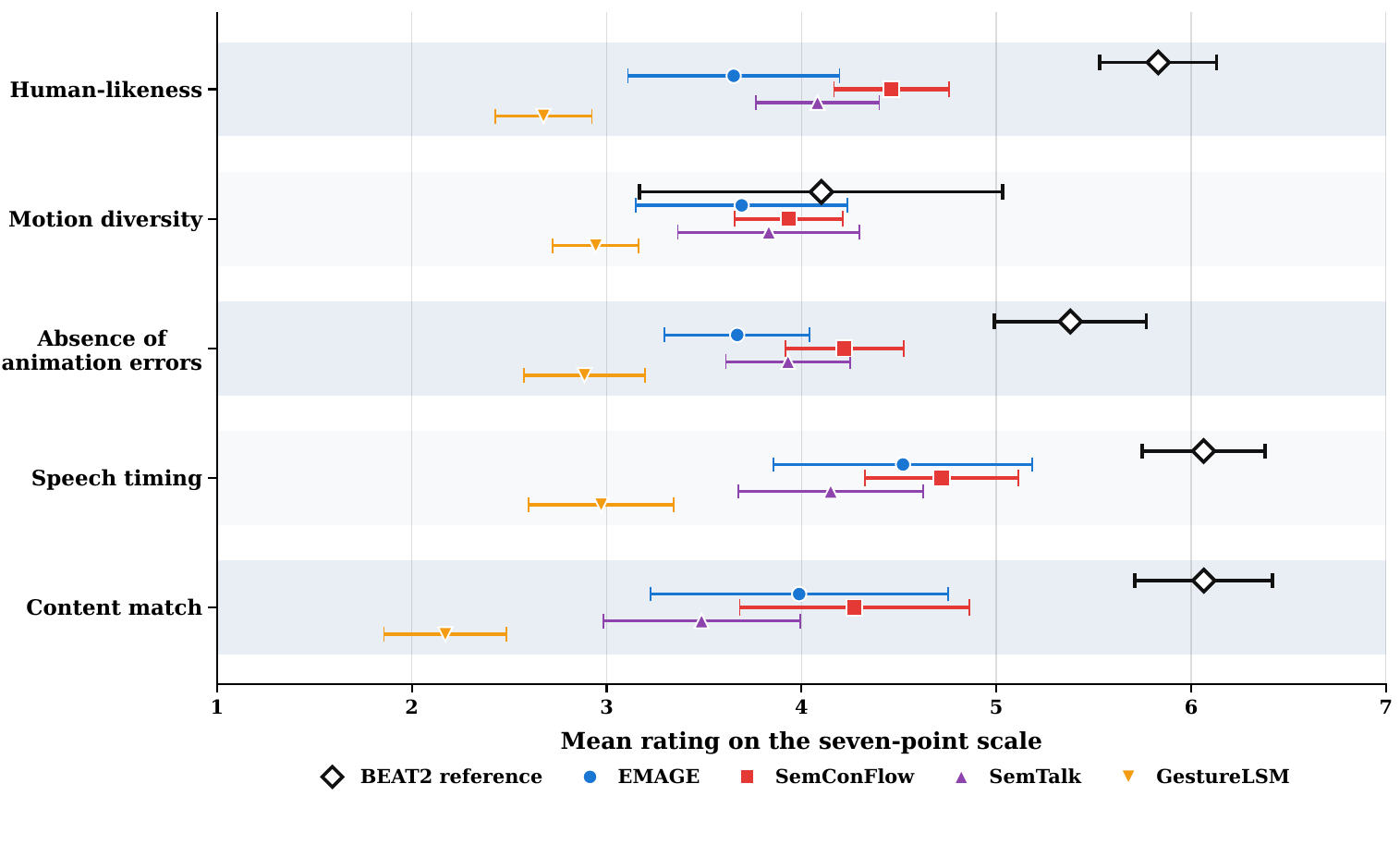}
    \caption{Subjective ratings across five perceptual dimensions. Points show mean ratings with standard-deviation error bars; BEAT2 serves as the reference, and the remaining methods are generated results.}
    \label{fig:human_study_results}
\end{figure}

The final subjective dataset contains ratings from 49 participants in the muted condition and 52 participants in the audio condition. Participants reported English fluency in both studies. The muted group had a mean age of 37.5 years ($\mathrm{SD}=10.9$; 24 male, 25 female), while the audio group had a mean age of 38.9 years ($\mathrm{SD}=11.9$; 29 male, 23 female). For each motion source, we first average ratings at the sequence level and then aggregate over the 15 unique sequences. Figure~\ref{fig:human_study_results} summarizes the resulting model means together with between-sequence variation on the common seven-point scale.

BEAT2 reference motion receives the highest mean rating across all five perceptual dimensions, with particularly large margins for human-likeness, absence of animation errors, speech timing, and content match. The gap is notably smaller for motion diversity, where the reference score is closer to the strongest generated result. This suggests that, under the predefined perceptual criterion of motion diversity used in our study, current systems can approach the reference more closely in perceived variation of motion, while a larger gap remains in overall human-likeness, animation quality, and speech-dependent appropriateness.

Among the generated systems, SemConFlow achieves the highest mean score on all five dimensions. In the muted condition, SemTalk generally ranks second, followed by EMAGE and GestureLSM, with SemTalk and EMAGE remaining relatively close on motion diversity. In the audio condition, EMAGE ranks second for both speech timing and content match, followed by SemTalk, while GestureLSM consistently receives the lowest mean ratings. The change in ordering between muted and audio evaluations indicates that visual motion quality and speech-conditioned appropriateness are not interchangeable perceptual properties.

Overall, the subjective results indicate that motion-centric and speech-aware criteria capture related but non-identical aspects of perceived gesture quality. Models do not preserve the same relative behavior across visual and speech-conditioned judgments, motivating an explicit analysis of how strongly the five perceptual dimensions covary before using them as separate targets for objective-metric validation.
\subsubsection{Relationship among perceptual dimensions.} 
\label{subsec: relamongperceptual}
To examine whether the five perceptual criteria capture distinct or shared aspects of gesture quality, we compute pairwise Spearman correlations between sequence-level subjective ratings. Table~\ref{tab:subjective_correlations} reports the associations for the 60 generated model--sequence observations, with the 15 BEAT2 reference sequences included as a complementary comparison. Among generated motions, all dimensions are positively correlated ($\rho=0.595$--$0.974$). Human-likeness is strongly associated with absence of animation errors ($\rho=0.928$), while motion diversity shows moderate-to-strong correlations with the other dimensions ($\rho=0.629$--$0.770$). Cross-condition correlations are also substantial, including human-likeness with speech timing ($\rho=0.650$) and content match ($\rho=0.624$). The strongest association is between speech timing and content match ($\rho=0.974$), indicating that temporal and semantic appropriateness are closely coupled in the evaluated stimuli. Overall, the results reveal shared perceptual aspects without making the five criteria interchangeable, supporting a multidimensional evaluation while highlighting the particularly close relationship between speech timing and content match.
\begin{table}[t]
\centering
\caption{Pairwise Spearman correlations among subjective perceptual dimensions.}
\label{tab:subjective_correlations}
\fontsize{6.8}{7.4}\selectfont
\setlength{\tabcolsep}{2.2pt}
\renewcommand{\arraystretch}{0.95}
\begin{tabular}{p{0.55\columnwidth}cc}
\toprule
\textbf{Perceptual dimension pair}
& \shortstack{\textbf{Generated}\\\textbf{($n=60$)}}
& \shortstack{\textbf{Including GT}\\\textbf{($n=75$)}} \\
\midrule
Human-likeness -- Motion diversity
& 0.728 & 0.626 \\
Human-likeness -- Absence of animation errors
& 0.928 & 0.960 \\
Human-likeness -- Speech timing
& 0.650 & 0.813 \\
Human-likeness -- Content match
& 0.624 & 0.801 \\
Motion diversity -- Absence of animation errors
& 0.770 & 0.647 \\
Motion diversity -- Speech timing
& 0.665 & 0.564 \\
Motion diversity -- Content match
& 0.629 & 0.538 \\
Absence of animation errors -- Speech timing
& 0.620 & 0.795 \\
Absence of animation errors -- Content match
& 0.595 & 0.783 \\
Speech timing -- Content match
& 0.974 & 0.985 \\
\bottomrule
\end{tabular}
\end{table}

\subsection{Objective–Subjective Correlation Analysis}
\begin{figure}[tbp]
    \centering
    \includegraphics[width=\linewidth]{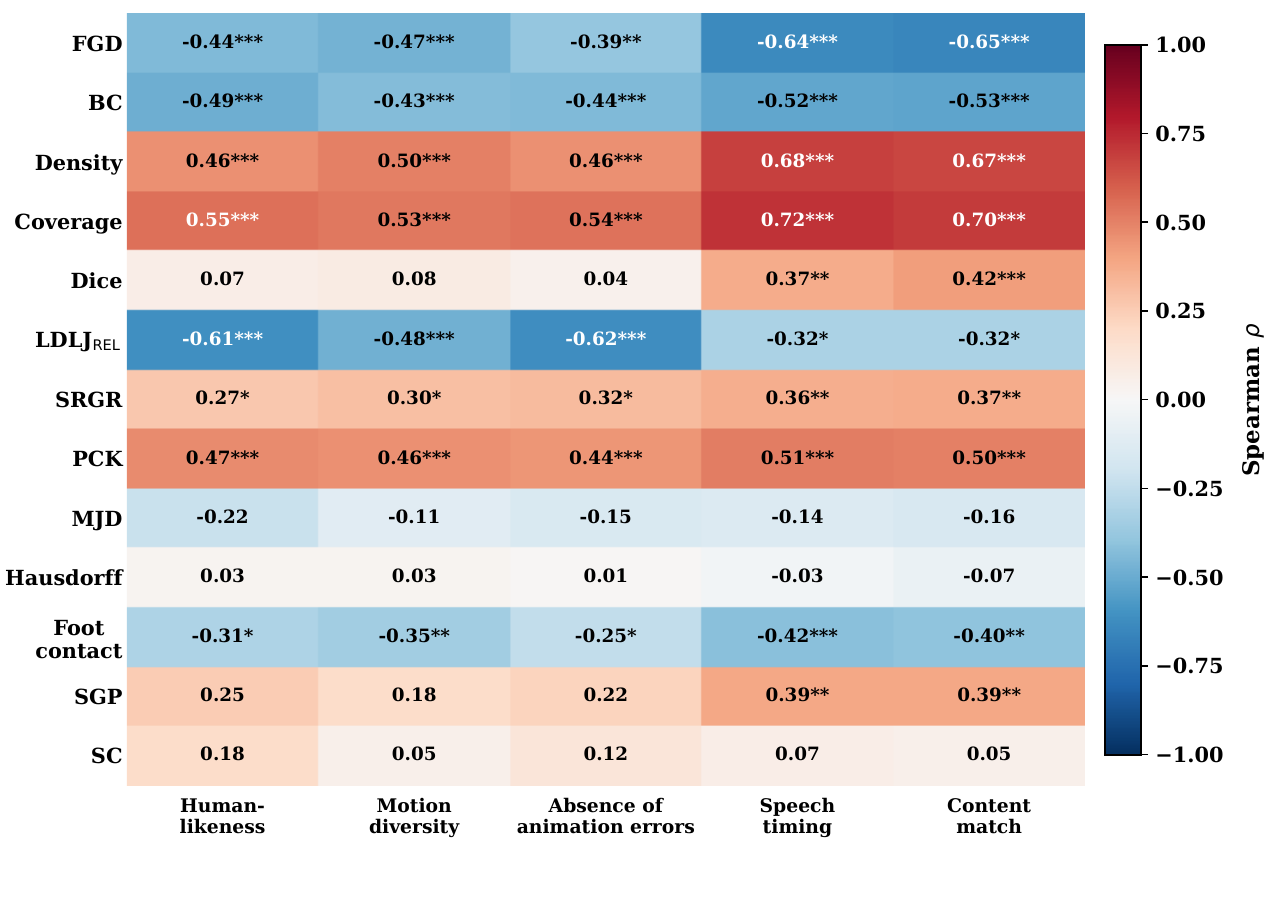}
    \caption{Spearman correlations between thirteen objective metrics and five subjective evaluation dimensions ($n=60$), with significance indicated by $^{*}p<0.05$, $^{**}p<0.01$, and $^{***}p<0.001$.}
    \label{fig:objective-subjective-correlations}
\end{figure}

Figure~\ref{fig:objective-subjective-correlations} shows a strongly dimension-dependent relationship between objective metrics and human judgments. No single metric consistently explains all five perceptual targets, indicating that objective measures capture distinct aspects of holistic gesture quality. SC is included in the analysis but does not exhibit a statistically meaningful correlation with any of the five subjective dimensions. For the muted dimensions, $\mathrm{LDLJ}_{REL}$ correlates most strongly with human-likeness ($\rho=-0.61$) and absence of animation errors ($\rho=-0.62$), while Coverage and Density are most informative for motion diversity. MJD and Hausdorff show little correspondence with subjective ratings, suggesting limited value of direct geometric distances for perceptual assessment.

For the speech-dependent dimensions, Coverage reaches $\rho=0.73$ for speech timing and $\rho=0.70$ for content match, with Density and FGD showing similarly strong associations. Although these metrics do not explicitly model semantics, their correlations indicate that motion-distribution quality can co-vary with speech appropriateness. BC, by contrast, is negatively correlated with all five perceptual dimensions, showing that stronger beat alignment alone does not imply better perceived quality.

The semantics-aware metrics provide an especially informative comparison. SC explicitly measures speech--motion similarity in a learned semantic embedding space, yet shows no statistically meaningful correlation with any of the five subjective dimensions, including speech timing and content match. In contrast, SGP exhibits the intended selectivity toward speech-aware judgments, correlating with both speech timing and content match at $\rho=0.39$, while remaining weak or non-significant for the muted dimensions. SGP also shows stronger alignment with human judgments on speech-aware dimensions than SRGR, while remaining comparable to SRGR on the other perceptual dimensions. Although its absolute correlations are lower than those of Coverage, Density, and FGD, SGP provides a more targeted and interpretable semantic signal by explicitly measuring whether semantic gesture intervals are preserved rather than reduced to generic beat-like motion. Together, the results for SC, SRGR, and SGP demonstrate that semantics-aware metrics differ substantially in their perceptual alignment and should therefore be validated directly against human judgments of communicative content. Overall, these findings motivate the subsequent composite analysis, where complementary objective signals are combined to better approximate human perception.

\subsection{Target-Specific Composite Metrics and Analysis}
\label{sec:composite_method}

\subsubsection{Learning Target-Specific Perceptual Composites}

Despite the positive associations among the subjective dimensions, the preceding analyses show that they are not interchangeable: their pairwise correlations vary substantially, model behavior differs across perceptual targets, and individual objective metrics exhibit strongly dimension-dependent correspondence with human judgments. We therefore construct a separate composite predictor for each perceptual dimension rather than collapsing gesture quality into a single global score.

Let \(\mathbf{x}_i=[x_{i1},\ldots,x_{iK}]^\top\), with \(K=13\), denote the objective-metric vector for model--sequence observation \(i\), comprising FGD, BC, Density, Coverage, Dice, \(\mathrm{LDLJ}_{\mathrm{REL}}\), SRGR, PCK, MJD, Hausdorff, Foot Contact, SC, and SGP. Chamfer Distance is excluded from the joint composite model because of its strong redundancy with Hausdorff Distance, while SC is retained as an independent semantics-aware predictor. Because these metrics operate on different numerical scales, each predictor is standardized using statistics estimated exclusively from the training observations,
\begin{equation}
z_{ik}
=
\frac{x_{ik}-\mu_k}{\sigma_k},
\label{eq:composite_standardization}
\end{equation}
where \(\mu_k\) and \(\sigma_k\) denote the training-set mean and standard deviation of metric \(k\).

For each perceptual target
\(q\in\{\mathrm{HL},\mathrm{MD},\mathrm{AE},\mathrm{ST},\mathrm{CM}\}\),
corresponding to human-likeness, motion diversity, absence of animation errors, speech timing, and content match, we fit an independent ordinary least-squares predictor with an intercept,
\begin{equation}
C_q(\mathbf{x}_i)
=
\beta_{0q}
+
\boldsymbol{\beta}_{q}^{\top}\mathbf{z}_i
=
\beta_{0q}
+
\sum_{k=1}^{K}\beta_{kq}z_{ik},
\label{eq:composite_metric}
\end{equation}
where \(\beta_{0q}\) is the intercept and \(\boldsymbol{\beta}_{q}\) contains the target-specific coefficients. The regression target is the mean subjective rating for dimension \(q\). Standardization is applied only to the predictors and does not normalize the subjective targets, so \(C_q(\mathbf{x}_i)\) remains on the original seven-point rating scale. For ordinary least squares with an intercept, fitting standardized predictors is algebraically equivalent to fitting the same predictors in their original units after reparameterizing the coefficients. Standardization therefore changes coefficient scaling but does not dampen or otherwise alter the fitted predictions.

To avoid information leakage, all standardization statistics and regression coefficients are estimated independently within each training fold. We evaluate the composite predictors using leave-one-out cross-validation (LOOCV) over the 60 generated model--sequence observations. In each fold, the held-out observation is standardized using statistics from the remaining 59 observations and predicted by the corresponding regression model, yielding one out-of-fold prediction per observation and perceptual target.

\subsubsection{Cross-Validated Perceptual Alignment}

Table~\ref{tab:composite_performance} evaluates how well each target-specific composite approximates its corresponding subjective dimension. Mean Squared Error (MSE) and Mean Absolute Error (MAE) quantify out-of-fold prediction error on the original seven-point scale, while Composite \(\rho\) denotes the Spearman correlation between the 60 out-of-fold predictions and subjective ratings. ``Best individual'' identifies the single objective metric with the strongest absolute Spearman correlation for that target, and
\begin{equation}
\Delta|\rho|
=
|\rho_{\mathrm{composite}}|
-
|\rho_{\mathrm{best\ individual}}|
\end{equation}
measures the gain over that metric.

Composite correlations range from \(0.5375\) for motion diversity to \(0.7824\) for content match. All five composites improve over their strongest individual metric, although the gains vary substantially across perceptual targets. The largest improvements are observed for absence of animation errors (\(+0.0819\)) and content match (\(+0.0805\)), followed by human-likeness (\(+0.0468\)). Motion diversity and speech timing show smaller gains of \(+0.0117\) and \(+0.0246\), respectively. These results indicate that the benefit of combining objective signals is target-dependent, with some perceptual dimensions drawing more strongly on complementary information distributed across multiple metrics.

\begin{table}[t]
\centering
\caption{
LOOCV performance of target-specific composite predictors over 60 generated model--sequence observations using 13 predictors. MSE and MAE report out-of-fold error on the seven-point subjective scale, while composite $\rho$ measures Spearman correlation with subjective ratings. Best individual is the strongest single objective metric by absolute Spearman correlation, and $\Delta|\rho|$ denotes the composite gain.
}
\label{tab:composite_performance}
\resizebox{\columnwidth}{!}{
\begin{tabular}{lccccc}
\toprule
Target Composite Metric
& MSE$\downarrow$
& MAE$\downarrow$
& Composite $\rho\uparrow$
& Best individual
& $\Delta|\rho|\uparrow$ \\
\midrule

Human-likeness
& 0.2846
& 0.4140
& 0.6568
& $\mathrm{LDLJ}_{\mathrm{REL}}$
& $+0.0468$ \\

Motion diversity
& 0.2035
& 0.3610
& 0.5375
& Coverage
& $+0.0117$ \\

Absence of animation errors
& 0.1560
& 0.2984
& 0.7065
& $\mathrm{LDLJ}_{\mathrm{REL}}$
& $+0.0819$ \\

Speech timing
& 0.2671
& 0.4349
& 0.7439
& Coverage
& $+0.0246$ \\

Content match
& 0.3665
& 0.4844
& 0.7824
& Coverage
& $+0.0805$ \\

\bottomrule
\end{tabular}
}
\end{table}

The composite predictors should be interpreted as dimension-specific automated proxies for human judgments rather than substitutes for direct perceptual evaluation. We therefore assess them at the sequence level through out-of-fold correlation and prediction error rather than deriving an additional model-level ranking from averaged predictions. The results show that the usefulness of metric combination is strongly target-dependent. The largest improvements occur for absence of animation errors and content match, suggesting that these perceptual judgments benefit from complementary signals across multiple objective measures. Human-likeness also shows a moderate improvement, whereas the gains for motion diversity and speech timing are comparatively small. Thus, no single pattern of metric combination applies uniformly across perceptual dimensions.

\section{Conclusions}

We present a perceptually grounded and semantics-aware benchmark for holistic co-speech gesture generation, combining standardized model comparison, human-centered metric validation, and fine-grained semantic evaluation. By retraining recent methods under a shared BEAT2 protocol and using a unified rendering and evaluation pipeline, we enable more controlled and directly comparable assessment across systems.

Our results show that gesture quality is strongly multidimensional. Model rankings vary across distributional, geometric, kinematic, temporal, and semantic measures, while no single objective metric consistently reflects all perceptual dimensions. Target-specific composite metrics improve alignment with human judgments across all five dimensions, with the largest gains observed for absence of animation errors and content match. These findings show that complementary objective signals can provide additional perceptual information beyond the strongest individual metric, although the benefit of combination remains target-dependent. 

For semantic evaluation, we enrich BEAT2 annotations with fine-grained gesture descriptions using Gemini 2.5 Pro and derive Semantic Gesture Preservation (SGP). Unlike SC, which shows no significant association with the evaluated perceptual dimensions, SGP is selectively aligned with speech-aware judgments and captures whether semantic gesture intervals are preserved rather than reduced to generic beat-like motion. Together, these findings provide a more systematic basis for evaluating progress in holistic co-speech gesture generation.
\section{Limitations and Discussion}
Although our benchmark reduces several sources of experimental variation, the evaluation remains constrained by the structure of the available data. The perceptual analysis is based on 15 unique speech sequences evaluated across four generation systems, yielding 60 generated model sequence observations rather than 60 fully independent samples. In addition, the selected study windows were deliberately enriched for semantic gestures, which improves the sensitivity of semantic evaluation but may overrepresent semantically dense speech relative to the broader BEAT2 distribution. 

The semantic analysis introduces a further limitation. Gemini~2.5~Pro provides a scalable way to enrich coarse BEAT annotations with gesture descriptions, contextual meaning, and gesture type, but its outputs should not be treated as semantic ground truth. Likewise, SGP specifically measures whether annotated semantic intervals avoid collapsing into generic beat gestures; it does not directly verify that the generated motion preserves the exact intended meaning or gesture category. More broadly, the observed objective--subjective relationships are specific to the evaluated models, sequences, and metric implementations. Future work should therefore extend the benchmark to larger and more diverse datasets, models, and perceptual studies, and develop semantic metrics that can more directly assess meaning preservation while retaining the scalability of multimodal-model-based evaluation.

\bibliographystyle{eg-alpha-doi} 
\bibliography{egbibsample}       


\newpage

\section{Appendix}
\subsection{User Study Interface and Instructions}
\label{sec:user_study_interface}

This section summarizes the participant-facing interfaces and instructions used in the two perceptual studies. The muted and audio conditions follow the same interaction flow but differ in the perceptual dimensions being assessed. In both studies, participants were required to watch each clip in full before submitting their ratings.

\begin{figure}[!ht]
    \centering
    \begin{subfigure}[t]{0.48\columnwidth}
        \centering
        \includegraphics[width=\linewidth]{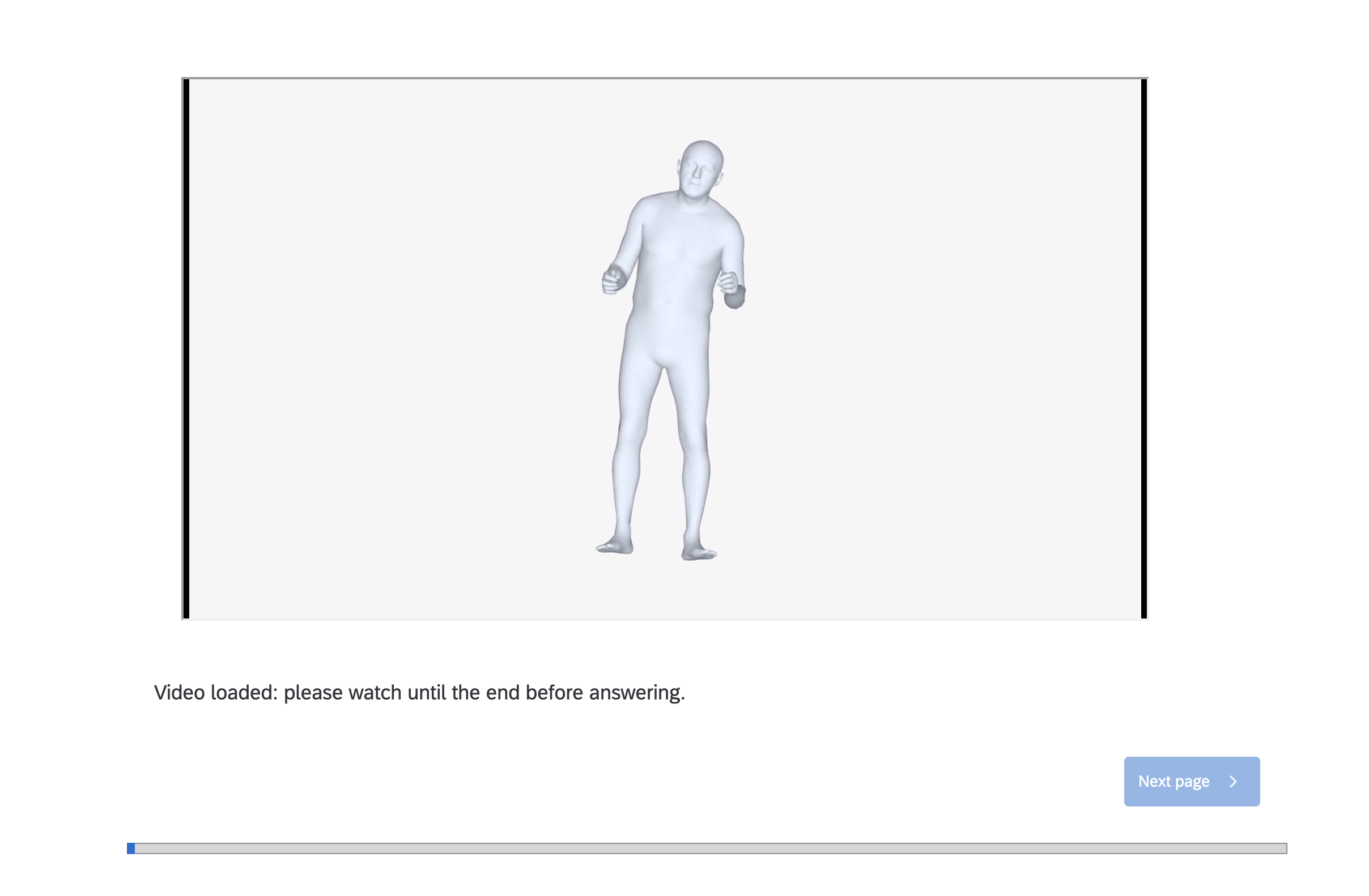}
        \caption{Before video playback.}
        \label{fig:muted_before}
    \end{subfigure}
    \hfill
    \begin{subfigure}[t]{0.48\columnwidth}
        \centering
        \includegraphics[width=\linewidth]{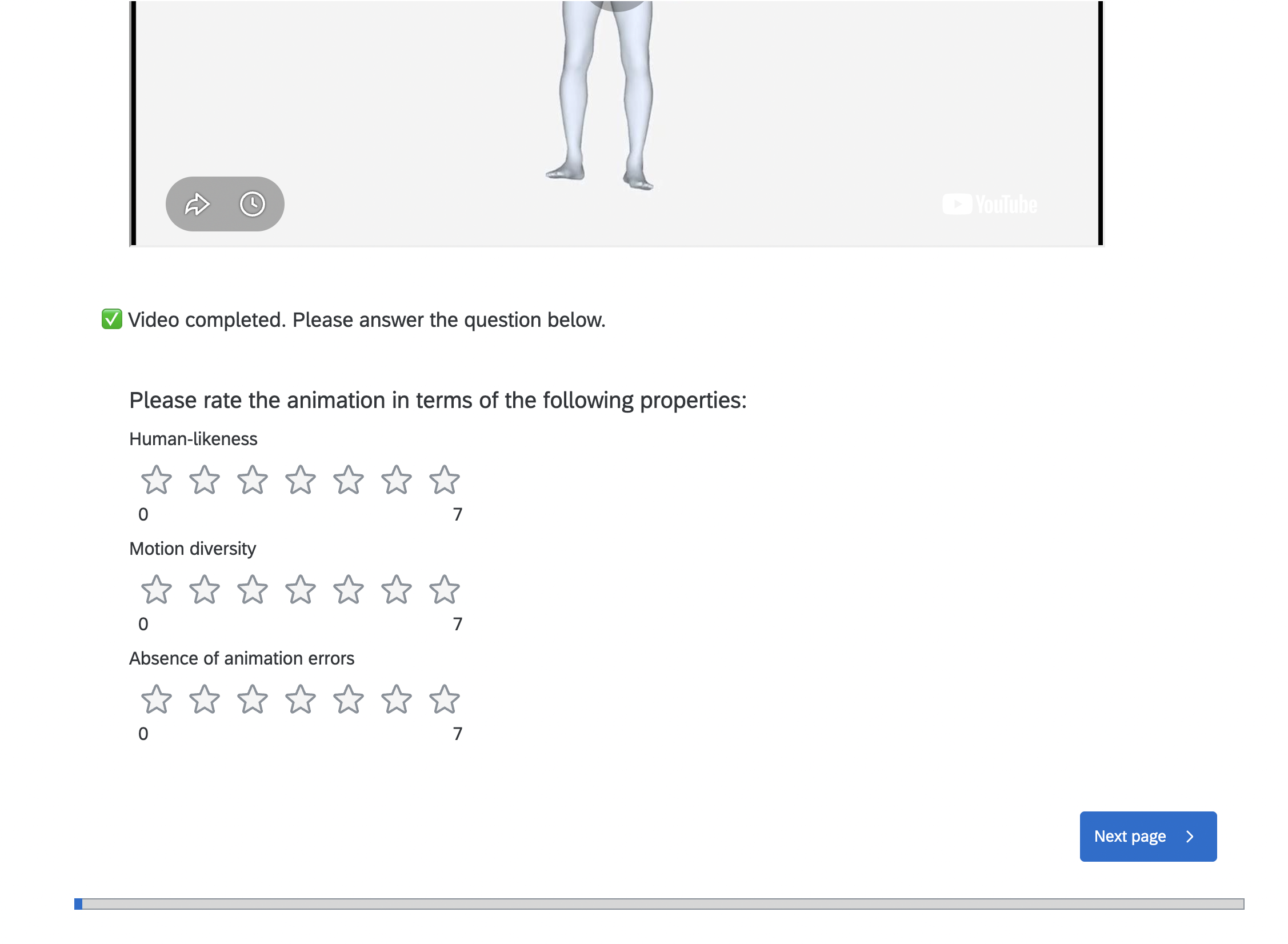}
        \caption{After video playback.}
        \label{fig:muted_after}
    \end{subfigure}
    \caption{
    Participant interface for the muted condition. Ratings become available only after the complete stimulus has been viewed.
    }
    \label{fig:muted_window}
\end{figure}

The muted condition focuses exclusively on visual motion quality. Participants first viewed the animation without audio and, once playback was completed, rated human-likeness, motion diversity, and absence of animation errors. Figure~\ref{fig:muted_window} illustrates the interface before and after video playback.

Before beginning the muted study, participants were given explicit definitions of the three evaluation criteria and instructed to base their judgments solely on the observed motion. The exact participant-facing instructions were as follows:

\begin{tcolorbox}[
    title=Participant Instructions: Muted Condition,
    colback=white,
    colframe=black!40,
    boxrule=0.5pt,
    arc=1mm,
    left=1.5mm,
    right=1.5mm,
    top=1mm,
    bottom=1mm
]
In the following pages, you will watch animations without audio and evaluate the body motion according to three criteria:

Human-likeness: Does the movement look like something a real person could perform?

Motion diversity: Does the animation contain a variety of different movements, rather than repeating the same motions throughout?

Absence of animation errors: Is the movement free from noticeable issues such as body parts suddenly jumping, getting stuck, moving through each other, or behaving strangely?

For each criterion, provide a rating on a 7-star scale, where 1 = Very poor and 7 = Excellent.

Please base your ratings only on the visual motion.

You will begin with 2 practice questions and 1 practice attention question.
\end{tcolorbox}

\begin{figure}[!ht]
    \centering
    \begin{subfigure}[t]{0.48\columnwidth}
        \centering
        \includegraphics[width=\linewidth]{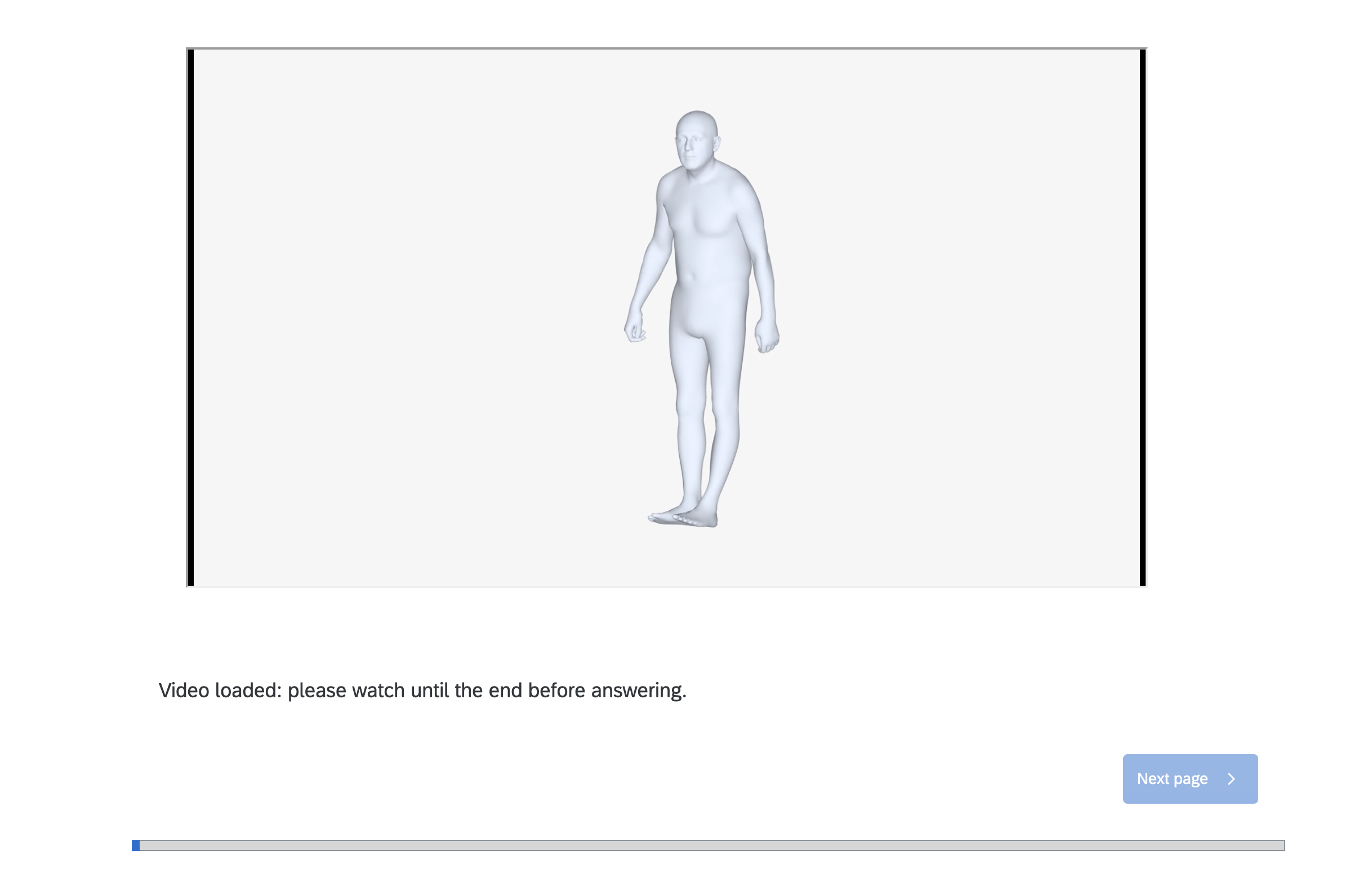}
        \caption{Before video playback.}
        \label{fig:audio_before}
    \end{subfigure}
    \hfill
    \begin{subfigure}[t]{0.48\columnwidth}
        \centering
        \includegraphics[width=\linewidth]{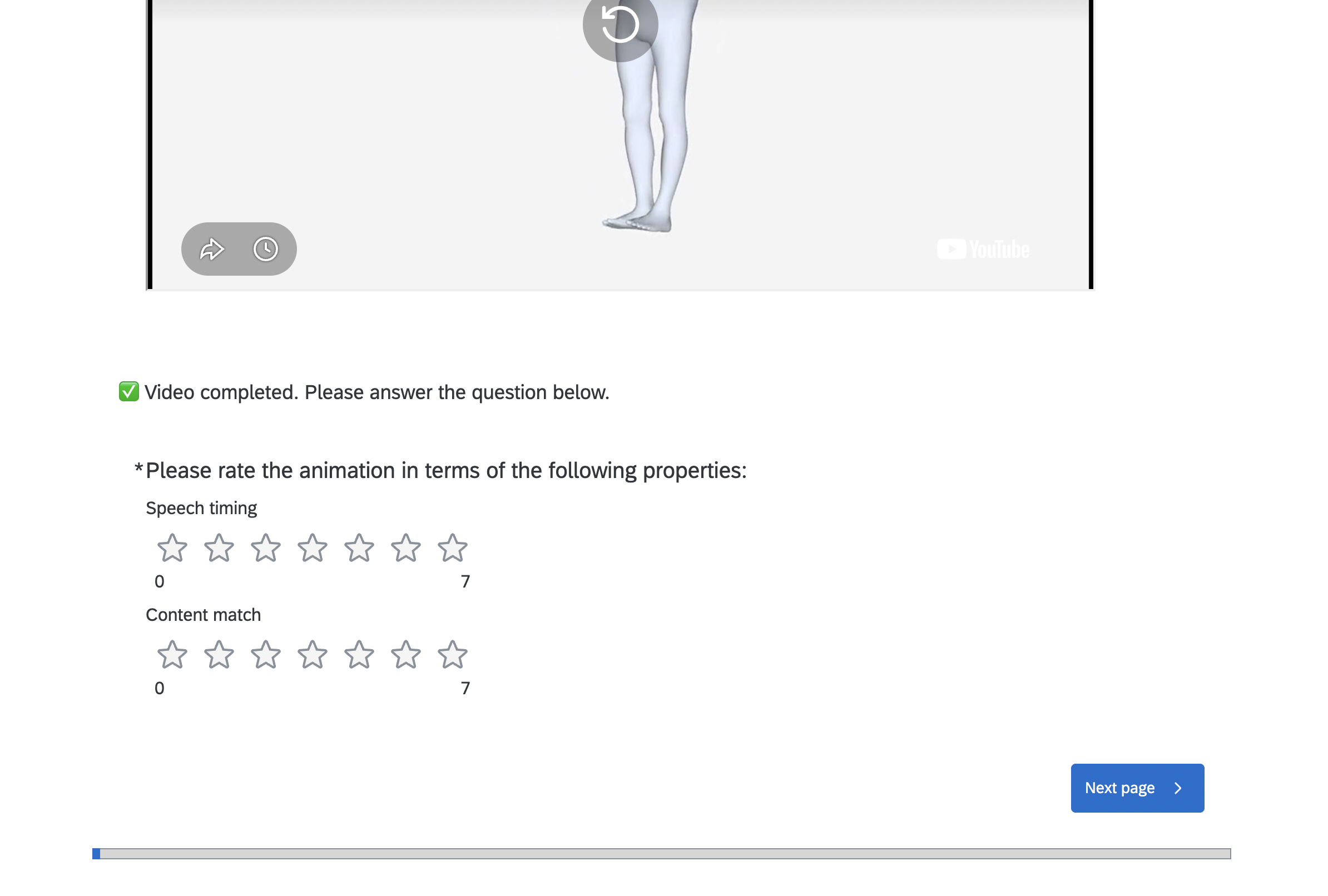}
        \caption{After video playback.}
        \label{fig:audio_after}
    \end{subfigure}
    \caption{
    Participant interface for the audio condition. After viewing the audiovisual stimulus, participants rate speech timing and content match.
    }
    \label{fig:audio_window}
\end{figure}

The audio condition follows the same presentation structure but evaluates speech-dependent properties. Participants viewed the animation together with the accompanying speech and subsequently rated speech timing and content match. The corresponding interface is shown in Figure~\ref{fig:audio_window}.

The instructions explicitly distinguish temporal coordination from semantic correspondence, allowing the two speech-dependent dimensions to be judged independently:

\begin{tcolorbox}[
    title=Participant Instructions: Audio Condition,
    colback=white,
    colframe=black!40,
    boxrule=0.5pt,
    arc=1mm,
    left=1.5mm,
    right=1.5mm,
    top=1mm,
    bottom=1mm
]
In the following pages, you will watch animations with audio and evaluate the gestures according to two criteria:

Speech timing: Do the gestures occur at appropriate moments relative to the rhythm, pauses, and emphasis of the speech?

Content match: Do the gestures meaningfully correspond to the spoken content?

For each criterion, provide a rating on a 7-star scale, where 1 = Very poor and 7 = Excellent.

Some attention-check questions regarding the spoken content will appear throughout the study. Please listen carefully before responding.

You will begin with 2 practice questions and 1 practice attention question.
\end{tcolorbox}

\begin{figure}[!ht]
    \centering
    \begin{subfigure}[t]{0.48\columnwidth}
        \centering
        \includegraphics[width=\linewidth]{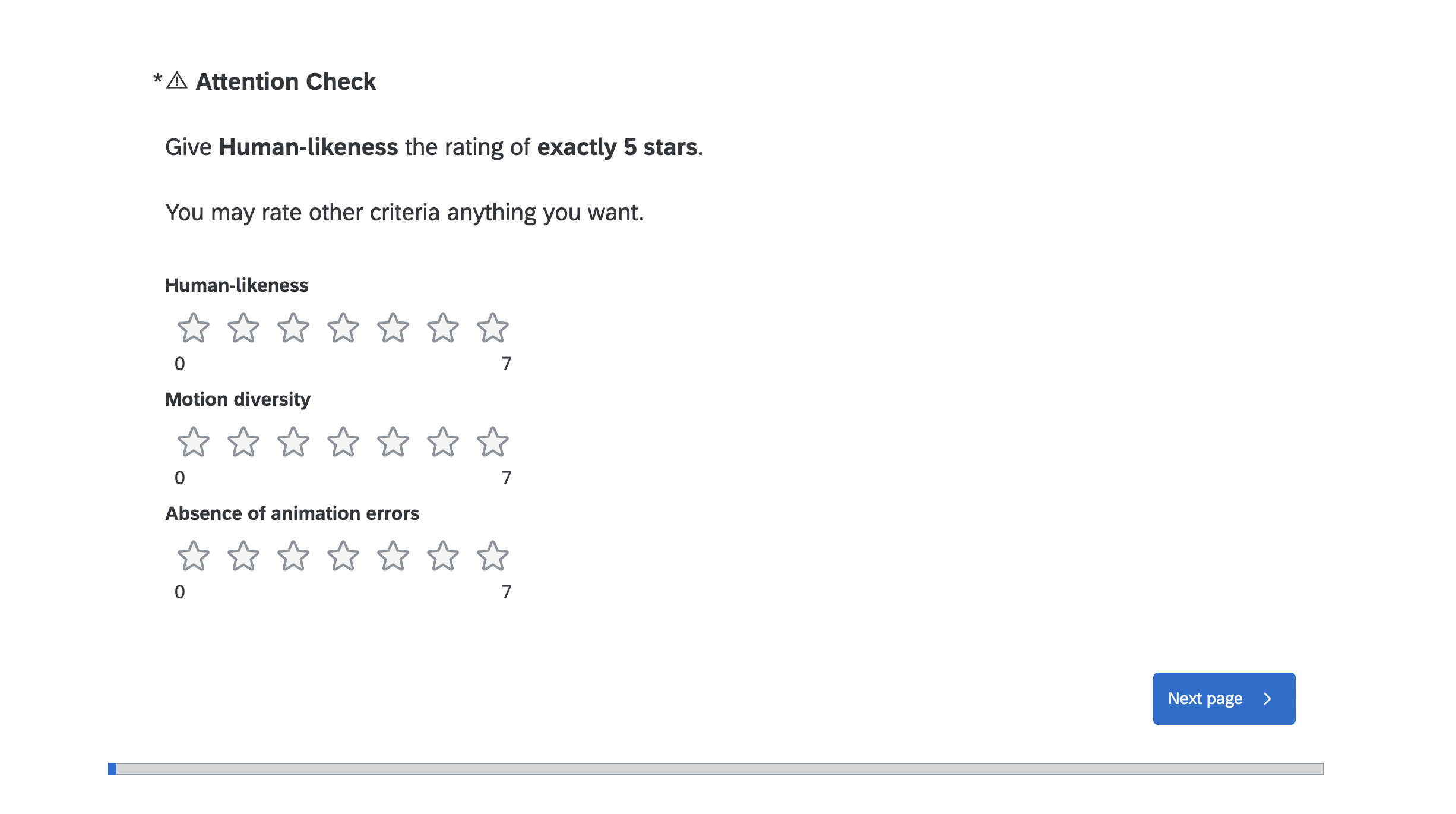}
        \caption{Muted condition.}
        \label{fig:attention_muted}
    \end{subfigure}
    \hfill
    \begin{subfigure}[t]{0.48\columnwidth}
        \centering
        \includegraphics[width=\linewidth]{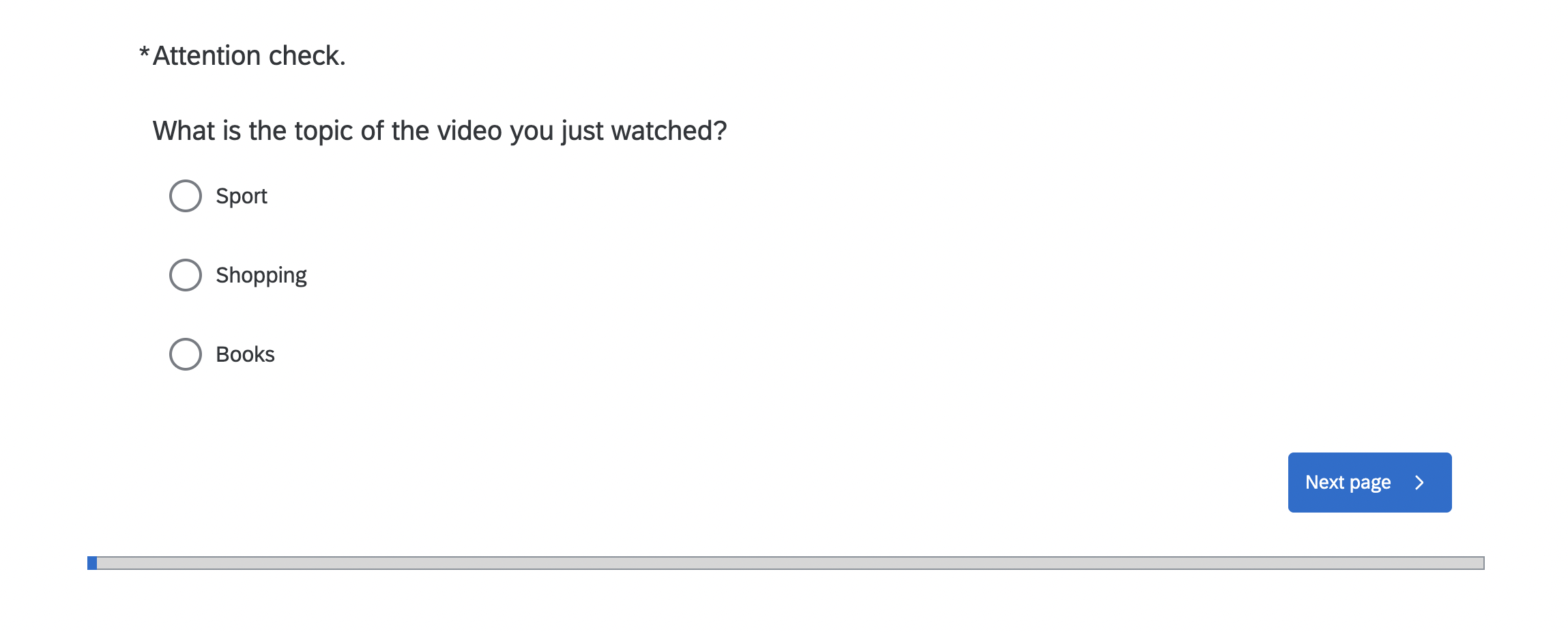}
        \caption{Audio condition.}
        \label{fig:attention_audio}
    \end{subfigure}
    \caption{
    Attention-check interfaces used in the two perceptual studies. The muted condition uses an instructed-response check, while the audio condition uses a content-based question about the preceding stimulus.
    }
    \label{fig:attention_questions}
\end{figure}

To monitor response quality, attention checks were distributed throughout both studies. The muted condition used instructed-response checks, whereas the audio condition used questions about the content of the preceding clip. Representative examples are shown in Figure~\ref{fig:attention_questions}.

\section{LLM-Based Semantic Gesture Annotation}
\label{app:semantic_annotation}

This section provides additional details of the LLM-based semantic gesture
annotation procedure introduced in Sec.~3.5. We describe the annotation
pipeline, the prompt used for multimodal annotation, an example output, and
human validation of the resulting annotations.

\subsection{Annotation Procedure}
\label{sec:annotation_procedure}

Our annotation pipeline converts predefined semantic gesture intervals into
fine-grained natural-language descriptions of gesture form and communicative
meaning. The procedure consists of four steps.

First, each source sequence is manually segmented into short video clips
according to the existing semantic-word intervals. Each clip therefore
corresponds to a predefined semantic expression and contains the motion
associated with that expression. Second, the semantic word associated with
each interval is retrieved from the existing annotation. Third, the segmented
video and its target semantic word are provided to Gemini using the prompt
shown in Fig.~\ref{fig:annotation_prompt}. The model is instructed to describe
only the gesture occurring within the specified interval and to characterize
its visible form in terms of handedness, hand shape, orientation, spatial
location, and movement trajectory. It additionally infers the communicative
meaning of the gesture and assigns one primary gesture class from
\{\textit{iconic}, \textit{metaphoric}, \textit{deictic}, \textit{beat}\}.
Finally, the generated annotations are checked and inserted into the
corresponding annotation files.

The resulting annotation for each semantic interval is represented as
\[
a_i =
(w_i, d_i, m_i, c_i),
\]
where $w_i$ denotes the target semantic word or expression, $d_i$ the
fine-grained gesture description, $m_i$ the inferred contextual meaning, and
$c_i$ the gesture class. This representation explicitly separates observable
motion form from its inferred semantic function, enabling subsequent
semantics-aware evaluation.

\begin{figure}[t]
\centering
\begin{tcolorbox}[
    width=\linewidth,
    colback=white,
    colframe=gray!70,
    boxrule=0.6pt,
    arc=1.5mm,
    left=2mm,
    right=2mm,
    top=1.5mm,
    bottom=2mm,
    title=\textbf{LLM Prompt: Semantic Gesture Annotation},
    colbacktitle=gray!65,
    coltitle=white,
    fonttitle=\bfseries\small
]
\scriptsize

You are an expert in multimodal gesture understanding.

\medskip
\textbf{TASK:}
Given a short video clip of a person speaking and the semantic-word interval
below, annotate only the gesture corresponding to the specified semantic word.

\medskip
\textbf{1. Gesture description.}
Describe the visible gesture involving the hands, arms, head, or upper body.
Specify:
(i) which hand(s) move;
(ii) hand shape;
(iii) palm or hand orientation;
(iv) spatial location relative to the body; and
(v) movement trajectory, direction, amplitude, and speed.
If multiple micro-gestures occur within the interval, describe them in
temporal order.

\medskip
\textbf{2. Contextual meaning.}
Infer the communicative function of the gesture in context.
Assign one primary class:
\textit{iconic}, \textit{metaphoric}, \textit{deictic}, or \textit{beat},
and briefly describe the conveyed meaning.

\medskip
\textbf{ANNOTATION SCOPE:}
Annotate only the specified semantic-word interval and ignore all other
time spans.

\medskip
\textbf{RULES:}
Be objective and concise.
Base the description strictly on what is visible within the target interval.
If the gesture is not visible or is ambiguous, use \texttt{unknown}.
If the hands are occluded, explicitly state this.
Do not borrow cues from outside the specified interval.
Prefer precise descriptions over speculative interpretations.

\medskip
\textbf{OUTPUT FORMAT:}

\medskip
{\ttfamily\scriptsize
\{
\par
\hspace*{1em}"word": "...",
\par
\hspace*{1em}"gesture\_description": "...",
\par
\hspace*{1em}"contextual\_meaning": "...",
\par
\hspace*{1em}"class": "...",
\par
\hspace*{1em}"confidence": ...
\par
\}
}

\end{tcolorbox}

\caption{\textbf{Prompt used for LLM-based semantic gesture annotation.}
The model receives a short video segment together with its predefined semantic
word and returns a structured description of gesture form and contextual
meaning.}
\label{fig:annotation_prompt}
\end{figure}

\subsection{Annotation Example}
\label{app:annotation_example}

\begin{figure}[t]
    \centering
    \includegraphics[width=0.2\linewidth]{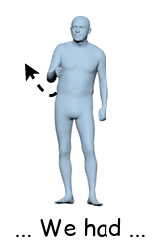}
    \caption{Representative event-level motion example used for semantic gesture annotation. The figure visualizes a short rendered motion clip of approximately 2--3\,s centered on the annotated semantic interval. This event clip, together with the corresponding target word and structured annotation prompt, is provided to Gemini~2.5~Pro as input for fine-grained semantic annotation.}
    \label{fig:event_video_example}
\end{figure}

In addition to the annotated intervals used throughout the dataset, we provide a representative event-level example in Figure~\ref{fig:event_video_example}. For this event, Gemini~2.5~Pro observes a rendered motion clip of approximately 2-3 seconds centered on the target semantic interval, together with the corresponding target word and the structured annotation prompt shown in ~\ref{fig:annotation_prompt}. The model is instructed to ground its response in the visible motion and to identify the observed gesture form, infer its contextual meaning, and assign a gesture type. Figure~\ref{fig:annotation_example} shows the resulting annotation output for the target word \textit{had}, illustrating how the proposed pipeline converts a short motion event into a structured semantic description.

\begin{figure}[t]
\centering
\begin{tcolorbox}[
    width=\linewidth,
    colback=white,
    colframe=gray!70,
    boxrule=0.6pt,
    arc=1.5mm,
    left=2mm,
    right=2mm,
    top=1.5mm,
    bottom=2mm,
    title=\textbf{Example annotation output for an additional semantic-gesture event},
    colbacktitle=gray!65,
    coltitle=white,
    fonttitle=\bfseries\small
]
\scriptsize

\textbf{Target semantic word:} \textit{had}

\medskip
\textbf{LLM annotation:}

\medskip
{\ttfamily\scriptsize
\{
\par
\hspace*{1em}"word": "had",
\par
\hspace*{1em}"gesture\_description":
\par
\hspace*{2em}"Right hand moves at shoulder level in front
\par
\hspace*{2em}of the torso. Motion is a small, straight
\par
\hspace*{2em}downward flick with very small amplitude
\par
\hspace*{2em}and speed.",
\par
\hspace*{1em}"contextual\_meaning":
\par
\hspace*{2em}"Holding or bringing an implied object into
\par
\hspace*{2em}possession.",
\par
\hspace*{1em}"class": "iconic",
\par
\hspace*{1em}"confidence": 0.58
\par
\}
}

\end{tcolorbox}

\caption{\textbf{Example of an LLM-generated semantic gesture annotation.}
The target semantic expression is paired with a fine-grained description of
the visible motion and an interpretation of its communicative meaning.}
\label{fig:annotation_example}
\end{figure}

\subsection{Human Validation}
\label{app:annotation_validation}

To assess the reliability of the automatically generated annotations, we
conducted a manual validation study on a randomly sampled subset of the
annotated semantic gestures. Human validation considered both the visual
description of the gesture and its inferred semantic interpretation.

For each sampled interval, the annotator viewed the corresponding video and
compared it with the LLM-generated annotation. Gesture descriptions were
checked with respect to five observable components:
\textit{handedness}, \textit{hand shape}, \textit{orientation},
\textit{location}, and \textit{movement}. The semantic interpretation was
additionally checked for whether the predicted gesture class and contextual
meaning were consistent with the observed gesture and associated semantic
expression.

An annotation was considered correct when its gesture description captured
the principal visible motion and its semantic interpretation was compatible
with the manually assessed communicative function. Under this criterion,
\textbf{75\%} of the evaluated LLM-generated annotations were judged correct.
This validation indicates that the LLM annotations provide a useful scalable
source of semantic supervision, while also motivating the use of human
verification for ambiguous gesture instances.

\end{document}